\documentclass[letterpaper, 10 pt, conference]{ieeeconf}  
\usepackage[noadjust]{cite}
\usepackage{amsmath}
\IEEEoverridecommandlockouts                              
\newcommand\T{\rule{0pt}{7.6ex}}       
\newcommand\B{\rule[-6.2ex]{0pt}{0pt}} 

\usepackage{etoolbox}
\makeatletter
\patchcmd{\@makecaption}
  {\scshape}
  {}
  {}
  {}
\makeatother

\usepackage{graphics} 
\usepackage{epsfig} 
\usepackage{amsmath} 
\usepackage{amssymb}  
\usepackage{multirow}
\usepackage{tabularx}
\usepackage{txfonts}

\title{\LARGE \bf
Detection and Segmentation of Manufacturing Defects with Convolutional Neural Networks and Transfer Learning
}

\author{Max Ferguson$^{1}$ Ronay Ak$^{2}$ Yung-Tsun Tina Lee$^{2}$ and Kincho. H. Law$^{1}$
\thanks{$^{1}$Stanford University, Civil and Environmental Engineering, CA, USA}%
\thanks{$^{2}$NIST, Systems Integration Division, Gaithersburg, MD, USA}%
}

\begin{document}

\maketitle
\thispagestyle{empty}
\pagestyle{empty}

\begin{abstract}

Quality control is a fundamental component of many manufacturing processes, especially those involving casting or welding. However, manual quality control procedures are often time-consuming and error-prone. In order to meet the growing demand for high-quality products, the use of intelligent visual inspection systems is becoming essential in production lines. Recently, Convolutional Neural Networks (CNNs) have shown outstanding performance in both image classification and localization tasks. In this article, a system is proposed for the identification of casting defects in X-ray images, based on the Mask Region-based CNN architecture. The proposed defect detection system simultaneously performs defect detection and segmentation on input images, making it suitable for a range of defect detection tasks. It is shown that training the network to simultaneously perform defect detection and defect instance segmentation, results in a higher defect detection accuracy than training on defect detection alone. Transfer learning is leveraged to reduce the training data demands and increase the prediction accuracy of the trained model. More specifically, the model is first trained with two large openly-available image datasets before finetuning on a relatively small metal casting X-ray dataset. The accuracy of the trained model exceeds state-of-the art performance on the GRIMA database of X-ray images  (GDXray) Castings dataset and is fast enough to be used in a production setting. The system also performs well on the GDXray Welds dataset. A number of in-depth studies are conducted to explore how transfer learning, multi-task learning, and multi-class learning influence the performance of the trained system.

\end{abstract}

\section{INTRODUCTION}

Quality management is a fundamental component of a manufacturing process \cite{rao_metal_2007}. To meet growth targets, manufacturers must increase their production rate while maintaining stringent quality control limits. In a recent report, the development of better quality management systems was described as the most important technology advancement for manufacturing business performance \cite{noauthor_future_2016}. In order to meet the growing demand for high-quality products, the use of intelligent visual inspection systems is becoming essential in production lines.

Processes such as casting and welding can introduce defects in the product which are detrimental to the final product quality 
\cite{rajkolhe_defects_2014}. Common casting defects include air holes, foreign-particle inclusions, shrinkage cavities, cracks, wrinkles, and casting fins \cite{li_improving_2006}. If undetected, these casting defects can lead to catastrophic failure of critical mechanical components, such as turbine blades, brake calipers, or vehicle driveshafts. Early detection of these defects can allow faulty products to be identified early in the manufacturing process, leading to time and cost savings \cite{ghorai_automatic_2013}. Automated quality control can be used to facilitate consistent and cost-effective inspection. The primary drivers for automated inspection systems include faster inspection rates, higher quality demands, and the need for more quantitative product evaluation that is not hampered by the effects of human fatigue.

Nondestructive evaluation techniques allow a product to be tested during the manufacturing process without jeopardizing the quality of the product. There are a number of nondestructive evaluation techniques available for producing two-dimensional and three-dimensional images of an object. Real-time X-ray imaging technology is widely used in defect detection systems in industry, such as on-line weld defect inspection \cite{ghorai_automatic_2013}. Ultrasonic inspection and magnetic particle inspection can also be used to measure the size and position of casting defects in cast components \cite{baillie_implementing_2007,lovejoy_magnetic_2012}. X-ray Computed Tomography (CT) can be used to visualize the internal structure of materials. Recent developments in high resolution X-ray computed tomography have made it possible to gain a three-dimensional characterization of porosity \cite{masad_characterization_2002}. However, automatically identifying casting defects in X-ray images still remains a challenging task in the automated inspection and computer vision domains.

\begin{figure}[thpb]
\centering
\includegraphics[width=0.46\textwidth]{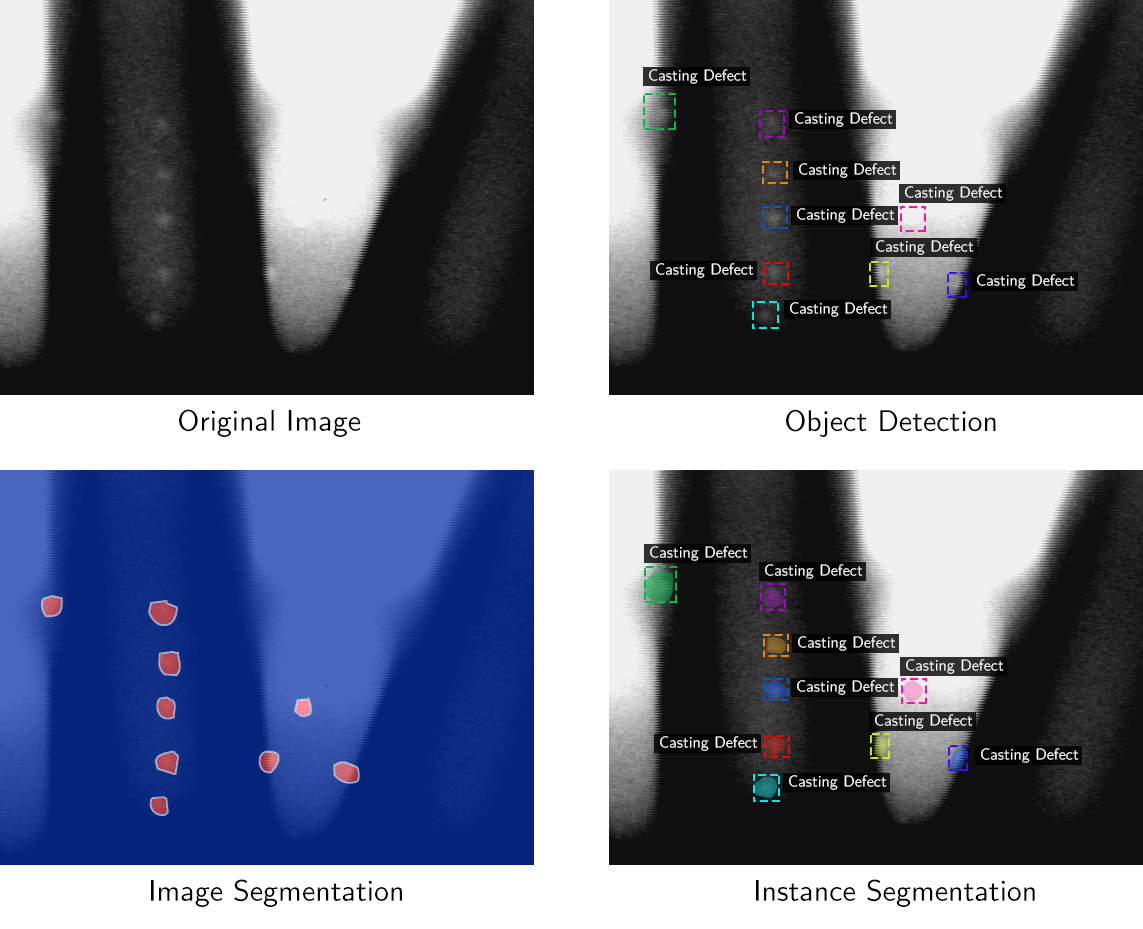}
\caption{Examples of different computer vision tasks for casting defect detection.}
\label{figure1}
\end{figure}

The defect detection process can be framed as either an object detection task or an instance segmentation task. In the object detection approach, the goal is to place a tight-fitting bounding box around each defect in the image. In the image segmentation approach, the problem is essentially one of pixel classification, where the goal is to classify each image pixel as a defect or not. Instance segmentation is a more difficult variant of image segmentation, where each segmented pixel must be assigned to a particular casting defect. A comparison of these computer vision tasks is provided in Figure \ref{figure1}. In general, object detection and instance segmentation are difficult tasks, as each object can cast an infinite number of different 2-D images onto the retina \cite{pinto_why_2008}. Additionally, the number of instances in a particular image is unknown and often unbounded. Variations of the object's position, pose, lighting, and background represent additional challenges to this task.

Many state-of-the-art object detection systems have been developed using the region-based convolutional neural network (R-CNN) architecture \cite{girshick_rich_2014}. R-CNN creates bounding boxes, or region proposals, using a process called selective search. At a high level, selective search looks at the image through windows of different sizes and, for each size, tries to group together adjacent pixels by texture, color, or intensity to identify objects. Once the proposals are created, R-CNN warps the region to a standard square size and passes it through a feature extractor. A support vector machine (SVM) classifier is then used to predict what object is present in the image, if any. In more recent object detection architectures, such as region-based fully convolutional networks (R-FCN), each component of the object detection network is replaced by a deep neural network \cite{dai_r-fcn:_2016}.

\begin{figure*}[thpb]
\centering
\includegraphics[width=0.9\textwidth]{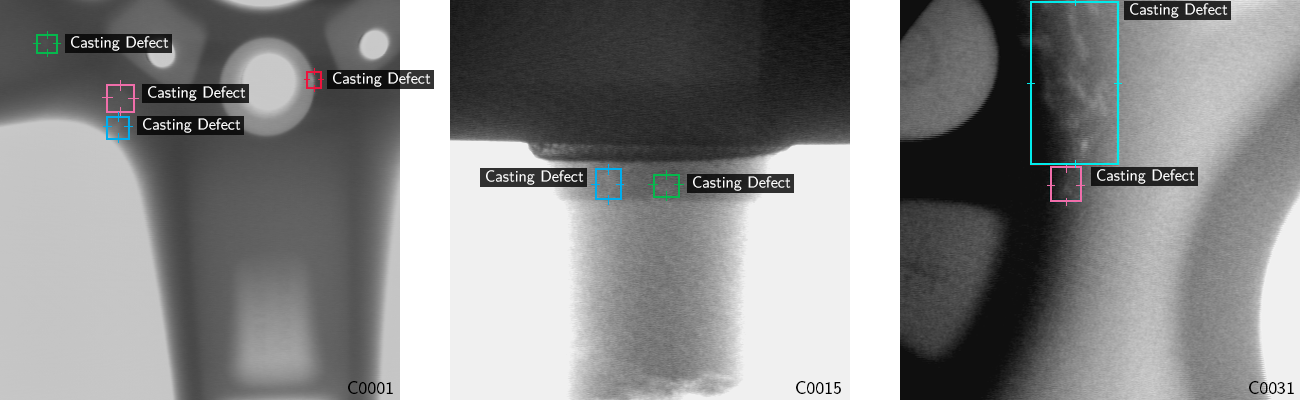}
\caption{Examples of X-ray images in the GDXray Castings dataset. The colored boxes show the ground-truth labels for casting defects.}
\label{figure2}
\end{figure*}

In this work, a fast and accurate defect detection system is developed by leveraging recent advances in computer vision. The proposed defect detection system is based on the mask region-based CNN (Mask R-CNN) architecture \cite{he_mask_2017}. This architecture simultaneously performs object detection and instance segmentation, making it useful for a range of automated inspection tasks. The proposed system is trained and evaluated on the GRIMA database of X-ray images (GDXray) dataset, published by Grupo de Inteligencia de Máquina (GRIMA) \cite{mery_gdxray:_2015}. Some examples from the GDXray dataset are shown in Figure \ref{figure2}.

The remainder of this article is organized as follows: The first section provides an overview of related works, and the second section provides a brief introduction to CNNs. A detailed description of the proposed defect detection system is provided in the “Defect Detection System” section. The “Implementation Details and Experimental Results” section explains how the system is trained to detect casting defects, and provides the main experimental results, as well as a comparison with similar systems in the literature. The article is concluded with a number of in-depth studies, a thorough discussion of the results, and a brief conclusion.

\section{RELATED WORKS}

The detection and segmentation of casting defects using traditional computer vision techniques has been relatively well-studied. One popular method is background subtraction, where an estimated background image (which does not contain the defects) is subtracted from the preprocessed image to leave a residual image containing the defects and random noise \cite{piccardi_background_2004,rebuffel_defect_2006}. Background subtraction has also been applied to the welding defect detection task, with varying levels of success \cite{nacereddine_weld_2005,wang_automatic_2002,kaftandjian_automatic_1998}. However, background subtraction tends to be very sensitive to the positioning of the image, as well as random image noise \cite{piccardi_background_2004}. A range of matched filtering techniques have also been proposed, with modified median (MODAN) filtering being a popular choice \cite{mery_review_2002}. The MODAN-Filter is a median filter with adapted filter masks, that is designed to differentiate structural contours of the casting piece from casting defects \cite{mackenzie_analytical_2005}. A number of researchers have proposed wavelet-based techniques with varying levels of success \cite{li_improving_2006,tang_application_2009}. In wavelet-based and frequency-based approaches, defects are commonly identified as high-frequency regions of the image, when compared to the comparatively lower frequency background \cite{zang_vision_2011}. Many of these approaches fail to combine local and global information from the image when classifying defects, making them unable to separate design features like holes and edges from casting defects.

In many traditional computer vision approaches, it is common to manually identify a number of features which can be used to classify individual pixels. Each image pixel is classified as a defect or treated as not being a defect, depending on the features that are computed from a local neighborhood around the pixel. Common features include statistical descriptors (mean, standard deviation, skewness, kurtosis) and localized wavelet decomposition \cite{li_improving_2006}. Several fuzzy logic approaches have also been proposed, but these techniques have been largely superseded by modern CNN-based computer vision techniques \cite{lashkia_defect_2001}.

The related task of automated surface inspection (ASI) is also well-documented in the literature. In ASI, surface defects are generally described as local anomalies in homogeneous textures. Depending on the properties of surface texture, ASI methods can be divided into four approaches \cite{xie_review_2008}. One approach is structural methods that model the texture primitives and displacements. Popular structural approaches include primitive measurement \cite{kittler_detection_1994}, edge features \cite{wen_verifying_1999}, and morphological operations \cite{mallik-goswami_detecting_2000}. The second approach is the statistical methods which measure the distribution of pixel values. The statistical approach is efficient for stochastic textures, such as ceramic tiles, castings, and wood. Popular statistical methods include histogram-based method \cite{kim_hierarchical_1994}, local binary pattern (LBP) \cite{niskanen_color_2001}, and co-occurrence matrix \cite{conners_identifying_1983}. The third approach is filter-based methods that apply filter banks on texture images. The filter-based methods can be divided into spatial-domain \cite{ade_comparison_1984}, frequency-domain \cite{hosseini_ravandi_fourier_1995}, and spatial-frequency domain \cite{hu_how_2016}. Finally, model-based approaches construct representations of images by modeling multiple properties of defects \cite{conci_fractal_1998}.

The research community, including this work, is greatly benefited from well-archived experimental datasets, such as the GRIMA database of X-ray images (GDXray) \cite{mery_gdxray:_2015}. The performance of several simple methods for defect segmentation are compared in \cite{mirzaei_automated_2017} using the GDXray Welds series, but each method  is only evaluated qualitatively. A comprehensive study of casting defect detection using various computer vision techniques is provided in \cite{mery_automatic_2017}, where patches of size $32 \times 32$ pixels are cropped from GDXray Castings series and used to train and test a number of different classifiers. The best performance is achieved by a simple LBP descriptor with a linear SVM classifier \cite{mery_automatic_2017}. Several deep learning approaches are also evaluated, obtaining up to 86.4 \% patch classification accuracy. When applying the deep learning techniques, the authors resize the 32 $\times$ 32 $\times$ 3 pixel patches to a size of  244 $\times$ 244 $\times$ 3 pixels so that they can be feed into pretrained neural networks \cite{mery_automatic_2017,szegedy_going_2015}. A deep CNN is used for weld defect segmentation in \cite{ren_experimental_2004} obtaining 90.5 \% accuracy on the binary classification of 25 $\times$ 25 pixel patches.

In recent times, a number of machine learning techniques have been successfully applied to the object detection task. Two notable neural network approaches are Faster Region-Based CNN (Faster R-CNN) \cite{ren_faster_2015} and Single Shot Multibox Detector (SSD) \cite{liu_ssd:_2016}. These approaches share many similarities, but the latter is designed to prioritize evaluation speed over accuracy. A comparison of different object detection networks is provided in \cite{huang_speed/accuracy_2017}. Mask R-CNN is an extension of Faster R-CNN that simultaneously performs object detection and instance segmentation \cite{he_mask_2017}. In previous research, it has been demonstrated that Faster R-CNN can be used as the basis for a fast and accurate defect detection system \cite{ferguson_automatic_2017}. This work builds on that progress by developing a defect detection system that simultaneously performs object detection and instance segmentation.

\section{CONVOLUTIONAL NEURAL NETWORKS}

\begin{figure*}[thpb]
\centering
\includegraphics[width=0.8\textwidth]{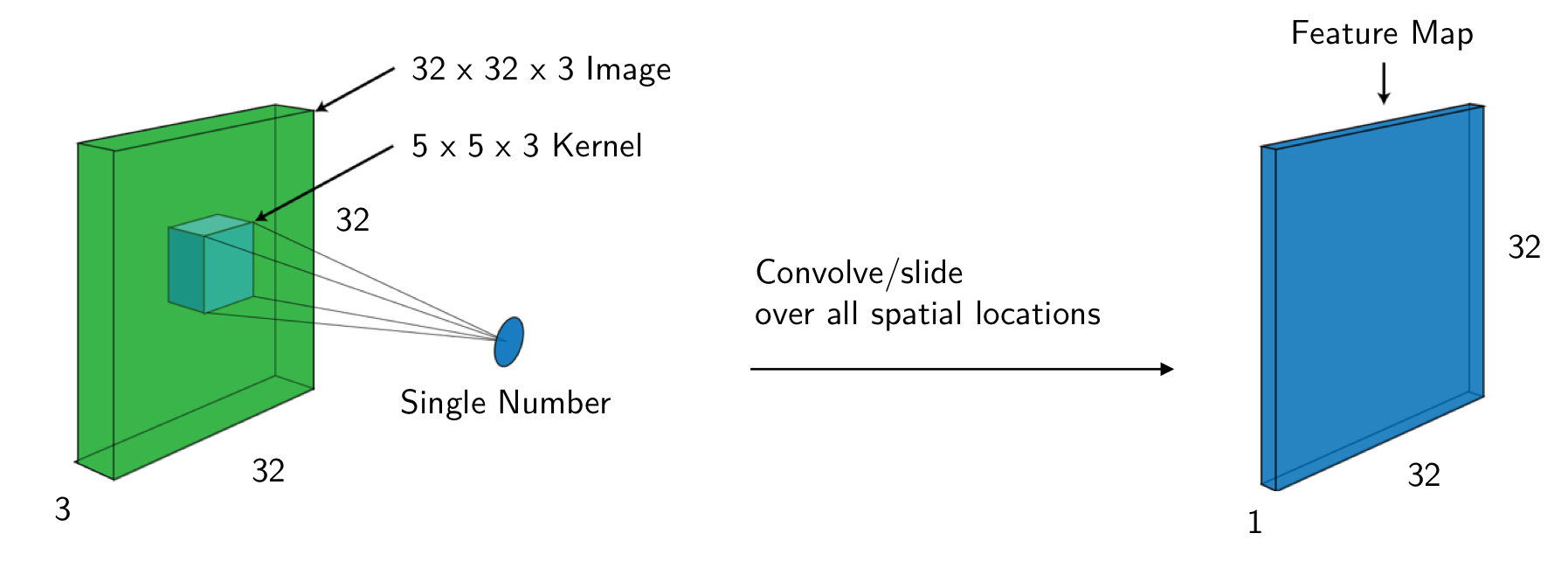}
\caption{Convolution of an image with a kernel to produce a feature map. Zero-padding is used to ensure that the spatial dimensions of the input layer are preserved. }
\label{figure3}
\end{figure*}

There has been significant progress in the field of computer vision, particularly in image classification, object detection and image segmentation. The development of deep CNNs has led to vast improvements in many image processing tasks. This section provides a brief overview of CNNs. For a more comprehensive description, the reader is referred to \cite{wu_convolutional_2017}.\\
In a CNN, pixels from each image are converted to a featurized representation through series of mathematical operations. Images can be represented as an order 3 tensor $\mathcal{I} \in \varmathbb{R}^{H \times W \times D}$ with height $H$, width $W$, and $D$ color channels \cite{wu_convolutional_2017}. The input sequentially goes through a number of processing steps, commonly referred to as layers. Each layer $i$, can be viewed as an arbitrary transformation $\boldsymbol{x}_{i+1} = f(\boldsymbol{x}_i; \boldsymbol{\theta}_i)$ with inputs $\boldsymbol{x}_i$, outputs $\boldsymbol{x}_{i+1}$, and parameters $\boldsymbol{\theta}_i$. The outputs of a layer are often referred to as a feature map. By combining multiple layers it is possible to develop a complex nonlinear function which can map high-dimensional data (such as images) to useful outputs (such as classification labels) \cite{wu_convolutional_2017}. More formally, a CNN can be thought of as the composition of number of functions:
\begin{align}
f(\boldsymbol{x})=f_N (f_2 (f_1 (\boldsymbol{x}_1;\boldsymbol{\theta}_1 );\boldsymbol{\theta}_2 )…);\boldsymbol{\theta}_N),
\end{align}

where $\boldsymbol{x}_1$ is the input to the CNN and $f(\boldsymbol{x})$ is the output. There are several layer types which are common to most modern CNNs, including convolution layers, pooling layers and batch normalization layers. A convolution layer is a function $f_i (\boldsymbol{x}_i;\boldsymbol{\theta}_i)$ that convolves one or more parameterized kernels with the input tensor, $\boldsymbol{x}_i$. Suppose the input $\boldsymbol{x}_i$  is an order 3 tensor with size $H_i \times W_i \times D_i$. A convolution kernel is also an order 3 tensor with size $H \times W \times D_i$. The kernel is convolved with the input by taking the dot product of the kernel with the input at each spatial location in the input. The convolution of a $H \times W \times 1$ kernel with an image is shown diagrammatically in Figure \ref{figure3}. By convolving certain types of kernels with the input image, it is possible to obtain meaningful outputs, such as the image gradients \cite{sobel_isotropic_1990}. In most modern CNN architectures, the first few convolutional layers extract features like edges and textures. Convolutional layers deeper in the network can extract features that span a greater spatial area of the image, such as object shapes.

Deep neural networks are, by design, parameterized nonlinear functions \cite{wu_convolutional_2017}. An activation function is applied to the output of a neural network layer to introduce this nonlinearity. Traditionally, the sigmoid function was used as the nonlinear activation function in neural networks. In modern architectures, the Rectified Linear Unit (ReLU) is more commonly used as the neuron activation function, as it performs best with respect to runtime and generalization error \cite{nair_rectified_2010}. The nonlinear ReLU function follows the formulation $f(\boldsymbol{z})= \max(0,\boldsymbol{z})$ for each value, $z$, in the input tensor $\boldsymbol{x}_i$. Unless otherwise specified, the ReLU activation function is used as the activation function in the defect detection system described in this article.\\
Pooling layers are also common in most modern CNN architectures \cite{wu_convolutional_2017,zhang_deep_2017}. The primary function of pooling layers is to progressively reduce the spatial size of the representation to reduce the number of parameters in the network, and hence control overfitting. Pooling layers typically apply a max or average operation over the spatial dimensions of the input tensor. The pooling operation is typically performed over a $2 \times 2$ or $3 \times 3$ area of the input tensor. By stacking pooling and convolutional layers, it is possible to build a network that allows a hierarchical and evolutionary development of raw pixel data towards powerful feature representations. \\
Training a neural network is performed by minimizing a loss function \cite{wu_convolutional_2017}. The loss function is normally a measure of the difference between the current output of the neural network and the ground truth. As long as each layer of the neural network is differentiable, it is possible to calculate the gradient of the loss function with respect to the parameters. The backpropagation algorithm allows the numerical gradients to be calculated efficiently \cite{werbos_backpropagation_1990}. A gradient-based optimization algorithm such as stochastic gradient descent (SGD) can be used to find the parameters that minimize the loss function.

\subsection{Residual Networks}

\begin{figure}[tb]
\centering
\includegraphics[width=0.4\textwidth]{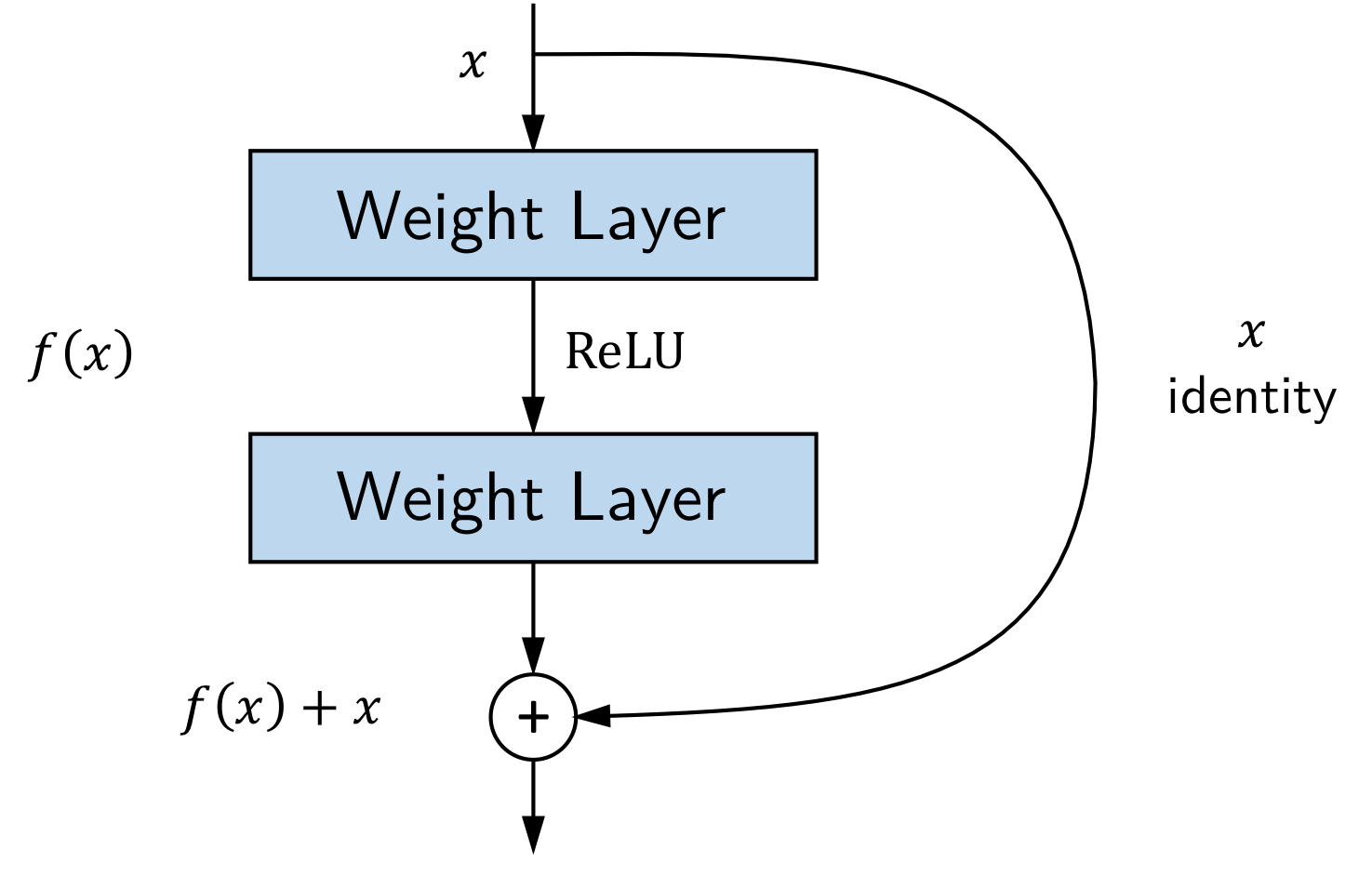}
\caption{A cell from the Residual Network architecture.}
\label{figure4}
\end{figure}

The properties of a neural network are characterized by choice and arrangement of the layers, often referred to as the architecture. Deeper networks generally allow more complex features to be computed from the input image. However, increasing the depth of a neural network often makes it more difficult to train, due to the vanishing gradient problem \cite{he_deep_2016}. The residual network (ResNet) architecture was designed to avoid many of the issues that plagued very deep neural networks. Most predominately, the use of residual connections helps to overcome the vanishing gradient problem \cite{he_deep_2016}. A cell from the ResNet architecture is shown in Figure \ref{figure4}. There are a number of standard variants of the ResNet architecture, containing between 18 and 152 layers. In this work, the relatively large ResNet-101 variant with 101 trainable layers is used as the neural network backbone \cite{he_deep_2016}. 

While ResNet was designed primarily to solve the image classification problem, it can also be used for a wider range of image processing tasks. More specifically, the outputs from the intermediate layers can be used as high-level representations of the image. When used this way, ResNet is referred to as a feature extractor, rather than a classification network.

\section{DEFECT DETECTION SYSTEM}
\begin{figure*}[tb]
\centering
\includegraphics[width=0.7\textwidth]{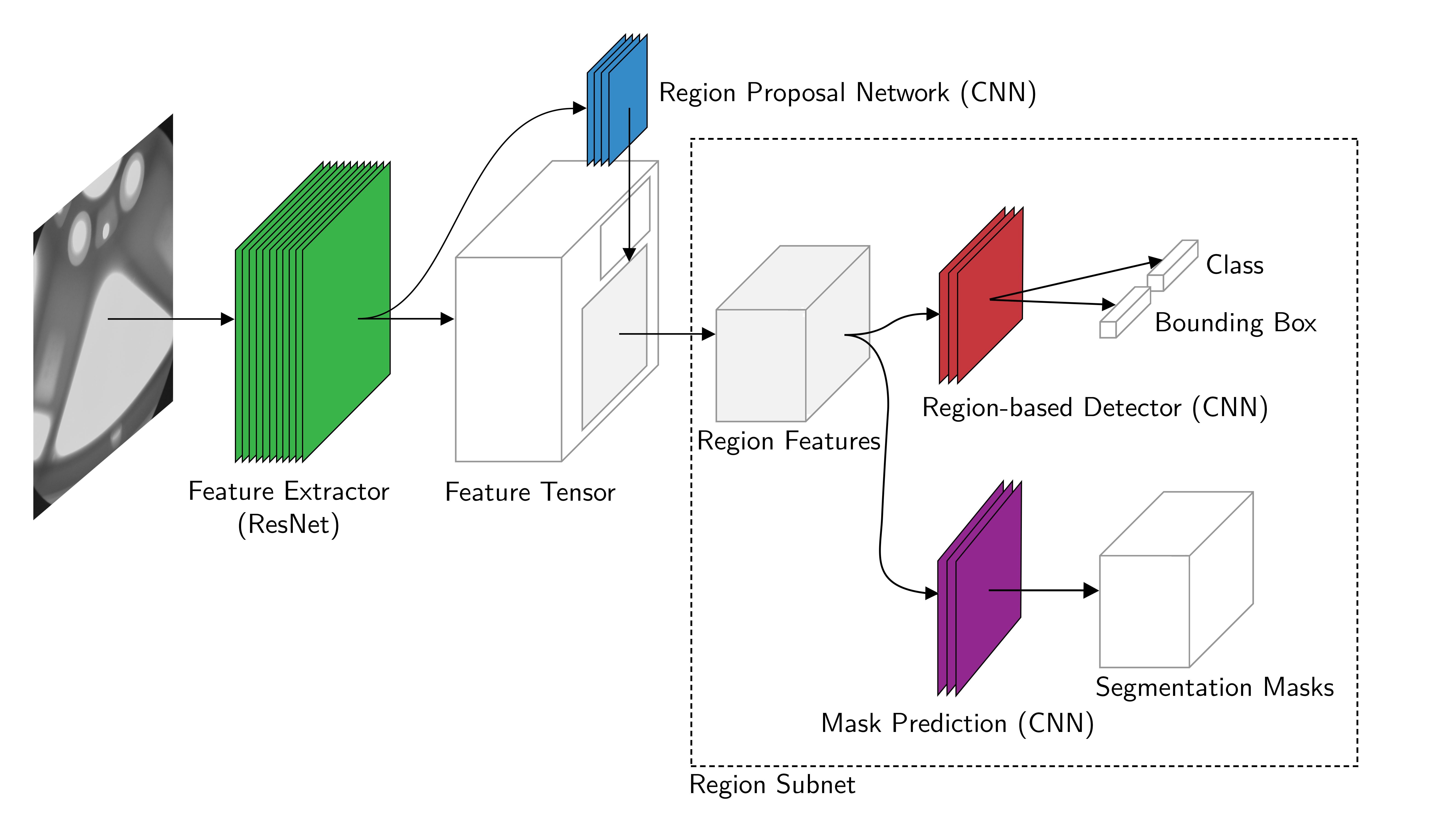}
\caption{The neural network architecture of the proposed defect detection system. The system consists of four convolutional neural networks, namely the ResNet-101 feature extractor, region proposal network, region-based detector and the mask prediction network.}
\label{figure5}
\end{figure*}

In this section, a defect detection system is proposed to identify casting defects in X-ray images. The proposed system simultaneously performs defect detection and defect segmentation, making it useful for a range of automated inspection applications. The design of the defect detection system is based on the Mask R-CNN architecture \cite{he_mask_2017}. As depicted in Figure \ref{figure5}, the defect detection system is composed of four modules. The first module is a feature extraction module that generates a high-level featurized representation of the input image. The second module is a CNN that proposes regions of interest (RoIs) in the image, based on the featurized image. The third module is a CNN that attempts to classify the objects in each RoI \cite{ren_faster_2015}. The fourth module performs image segmentation, with the goal of generating a binary mask for each region. Each module is described in detail throughout the remainder of this section. 

\subsection{Feature Extraction}

\begin{figure}[tb]
\centering
\includegraphics[width=0.90\linewidth]{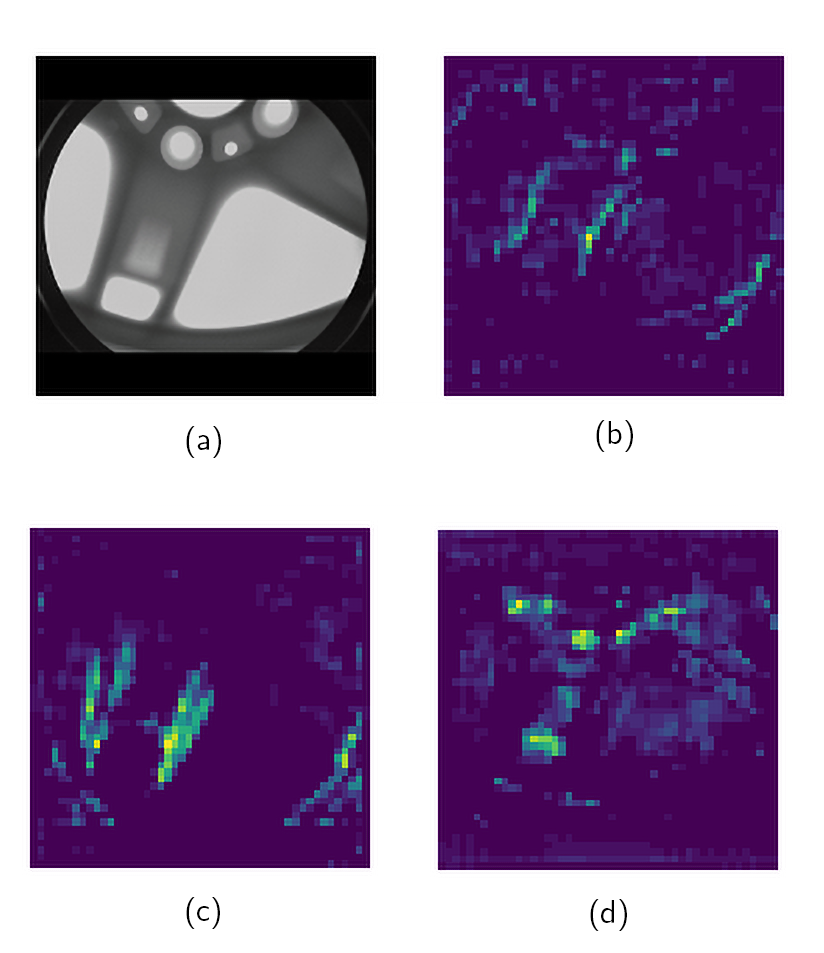}
\caption{Feature maps from the last layer of the "conv 4" ResNet feature extractor. Clockwise from the top left image: (a) the resized and padded X-ray image, (b) a feature map which appears to capture horizontal gradients (c) a feature map which appears to capture long straight vertical edges, (d) a feature map which appears to capture hole-shaped objects.}
\label{figure6}
\end{figure}

The first module in the proposed defect detection system transforms the image pixels into a high-level featurized representation. Many CNN-based object detection systems use the VGG-16 architecture to extract features from the input image \cite{girshick_rich_2014,ren_faster_2015,girshick_fast_2015}. However, recent work has demonstrated that better results can be obtained with more modern feature extractors \cite{huang_speed/accuracy_2017}. In a related work, we have shown that an object detection network with the ResNet-101 feature extractor results in a higher bounding-box prediction accuracy on the GDXray Castings dataset, than the same object detection network with a VGG-16 feature extractor \cite{ferguson_automatic_2017}. Therefore, the ResNet-101 architecture is chosen as the backbone for the feature extraction module. The neural-network architecture of the feature extraction module is shown in Table 1. Some feature maps from the feature extraction module are shown in Figure \ref{figure6}. 

The ResNet-101 feature extractor is a very deep convolutional neural network with 101 trainable layers and approximately 27 million parameters. Hence, it is unlikely that the network can be trained to extract meaningful features from input images, using the relatively small GDXray dataset.  One interesting property of CNN-based feature extractors is that the features they generate often transfer well across different image processing tasks. This property is leveraged when training the proposed casting defect detection system, by first training the feature extractor on the large ImageNet dataset \cite{russakovsky_imagenet_2015}. Throughout the training process the feature extractor learns to extract many different types of features, only some of which are useful on the comparatively simpler casting defect detection task. When training the object detection network on the GDXray Castings dataset, the system learns which features correlate well with casting defects and discards unneeded features. This process tends to work well, as it is much easier to discard unneeded features than it is to learn entirely new features. 

\begin{table}[h]
\caption{The neural network architecture used for feature extraction. The architecture is based on the ResNet-101 architecture, but excludes the "conv5\_x" block which is primarily designed for image classification \cite{zhang_deep_2017}. The term stride refers to the step size of the convolution operation. }
\label{resnet}
\begin{center}
\setlength{\tabcolsep}{10pt}
\renewcommand{\arraystretch}{1.5}
\begin{tabular}{|c|c|c|}
\hline
\textbf{Layer Name} & \textbf{Filter Size} & \textbf{Output Size}\\
\hline
conv1 & $7 \times 7, 64, \text{stride }2$ &  $ 112 \times 112 \times 64 $\\
\hline
conv2\_x &  
\T
$\begin{bmatrix}
    {1\times 1, 64} \\
    {3\times 3, 64} \\
    {1\times 1, 256} \\
\end{bmatrix}
\times 3$ \B
&  $ 112 \times 112 \times 64 $\\
\hline
conv3\_x & 
\T
$\begin{bmatrix}
    {1\times 1, 128} \\
    {3\times 3, 128} \\
    {1\times 1, 512} \\
\end{bmatrix} \times 4$
\B
&  $ 112 \times 112 \times 64 $\\
\hline
conv4\_x & $
\T
\begin{bmatrix}
    {1\times 1, 256} \\
    {3\times 3, 256} \\
    {1\times 1, 1024} \\
\end{bmatrix}\times 23$
 \B &  $ 112 \times 112 \times 64 $\\
\hline
\end{tabular}
\end{center}
\end{table}

\subsection{Region Proposal Network}

\begin{figure}[tb]
\centering
\includegraphics[width=0.3\textwidth]{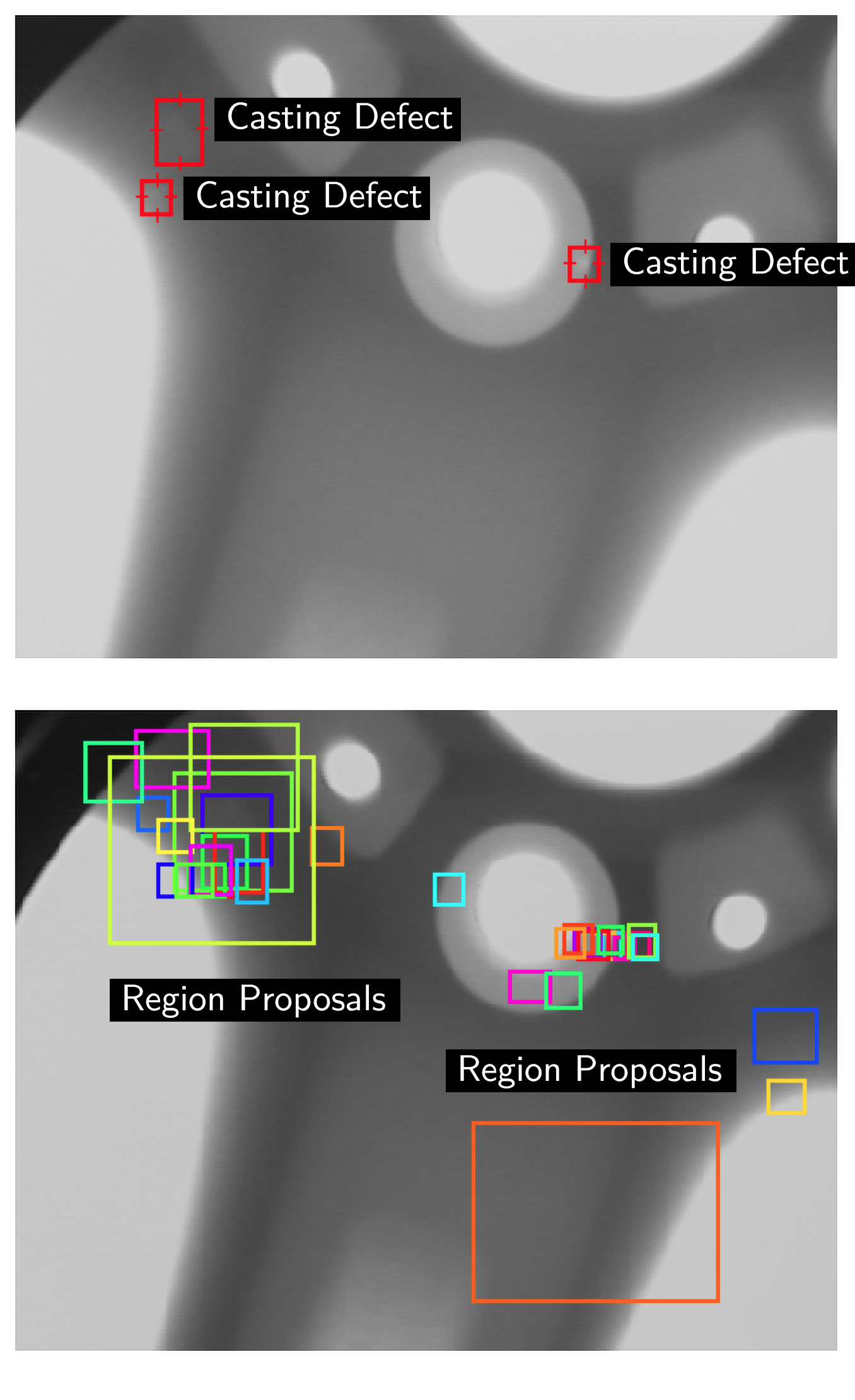}
\caption{Ground truth casting defect locations (top). The top 50 region proposals from the RPN for the same X-Ray image (bottom).}
\label{figure7}
\end{figure}

The second module in the proposed defect detection system is the region proposal network (RPN). The RPN takes a feature map of any size as input and outputs a set of rectangular object proposals, each with a score describing the likelihood that the region contains an object. To generate region proposals, a small CNN is convolved with the output of the ResNet-101 feature extractor. The input to this small CNN is an $n \times n$ spatial window of the ResNet-101 feature map. At a high-level, the output of the RPN is a vector describing the bounding box coordinates and likeliness of objects at the current sliding position. An example output containing 50 region proposals is shown in Figure \ref{figure7}. \\

\begin{figure}[thb]
\centering
\includegraphics[width=0.36\textwidth]{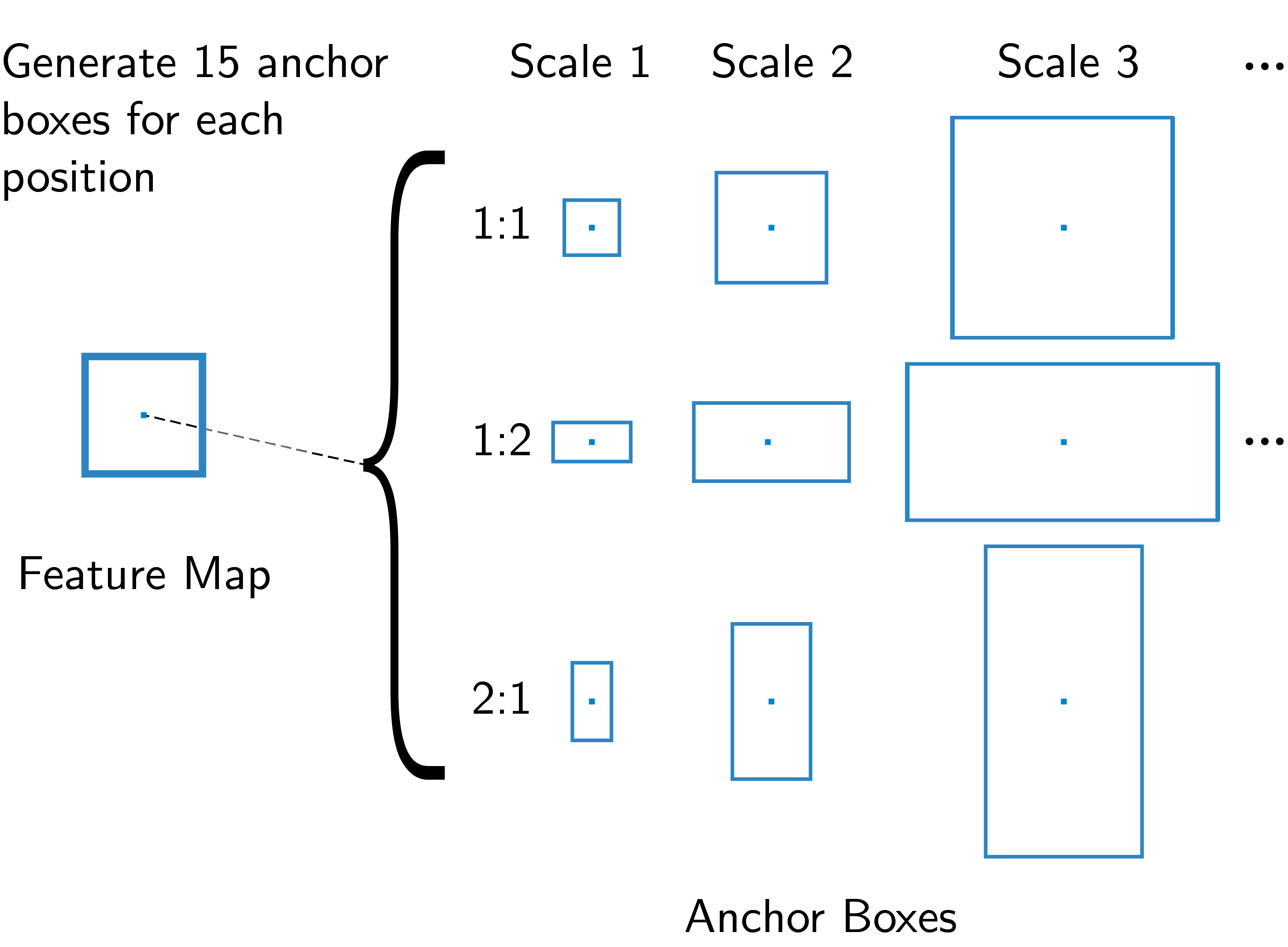}
\caption{Anchor Boxes at a certain position in the feature map.}
\label{figure8}
\end{figure}

\textbf{Anchor Boxes}: Casting defects come in a range of different scales and aspect ratios. To accurately identify casting defects, it is necessary to evaluate boxes with a range of box shapes, at every location in the image. These boxes are commonly referred to as anchor boxes. Anchors vary in aspect-ratio and scale, so as to contain any potential object in the image. At each sliding location, the RPN estimates the likelihood that each anchor box contains an object. The anchor boxes for one position in the feature map are shown in Figure \ref{figure8}. In this work, anchor boxes with 3 scales and 5 aspect ratios are used, yielding 15 anchors at each sliding position. The total number of anchors in each image depends on the size of the image. For a convolutional feature map of a size $W \times H$ (typically $\sim$42,400), there are $15WH$ anchors in total.

The size and scale of the anchor boxes are chosen to match the size and scale of objects in the dataset. It is common to use anchor boxes with areas of 1282, 2562, and 5122 pixels and aspect ratios of 1:1, 1:2, and 2:1, for detection of common objects like people and cars \cite{ren_faster_2015}. However, many of the casting defects in the GDXray dataset are on the scale of $20 \times 20$ pixels. Therefore, the smallest anchor box is chosen to be $16 \times 16$ pixels. Aspect ratios 1:1, 1:2, and 2:1 are used. Scale factors of 1, 2, 4, 8, and 16 are used. Most defects in the dataset are smaller than $64 \times 64$ pixels, so using scales 1, 2, and 4 could be considered sufficient for the defect detection task. However, the object detection network is pretrained on a dataset with many large objects, so the larger scales are included to avoid restricting the system during the pretraining phase. 

\textbf{Architecture}: The RPN predicts the bounding box coordinates and probability that the box contains an object, for all $k$ anchor boxes at each sliding position. The n × n input from the feature extractor is first mapped to a lower-dimensional feature vector (512-d) using a fully connected neural network layer. This feature vector is fed into two sibling fully-connected layers: a box-regression layer ($loc$) and a box-classification layer ($cls$). The class layer outputs $2k$ scores that estimate the probability of object and not object for each anchor box. The loc layer has $4k$ outputs, which encode the coordinate adjustments for each of the $k$ boxes. The reader is referred to \cite{ren_faster_2015} for a detailed description of the neural network architecture. The probability that an anchor box contains an object is referred to as the objectness score of the anchor box. This objectness score can be thought of as a way to distinguish objects in the image from the background. At the end of the region proposal stage, the top n anchor boxes are selected by objectness score as the region proposals.

\begin{figure}[tb]
\centering
\includegraphics[width=0.50\textwidth]{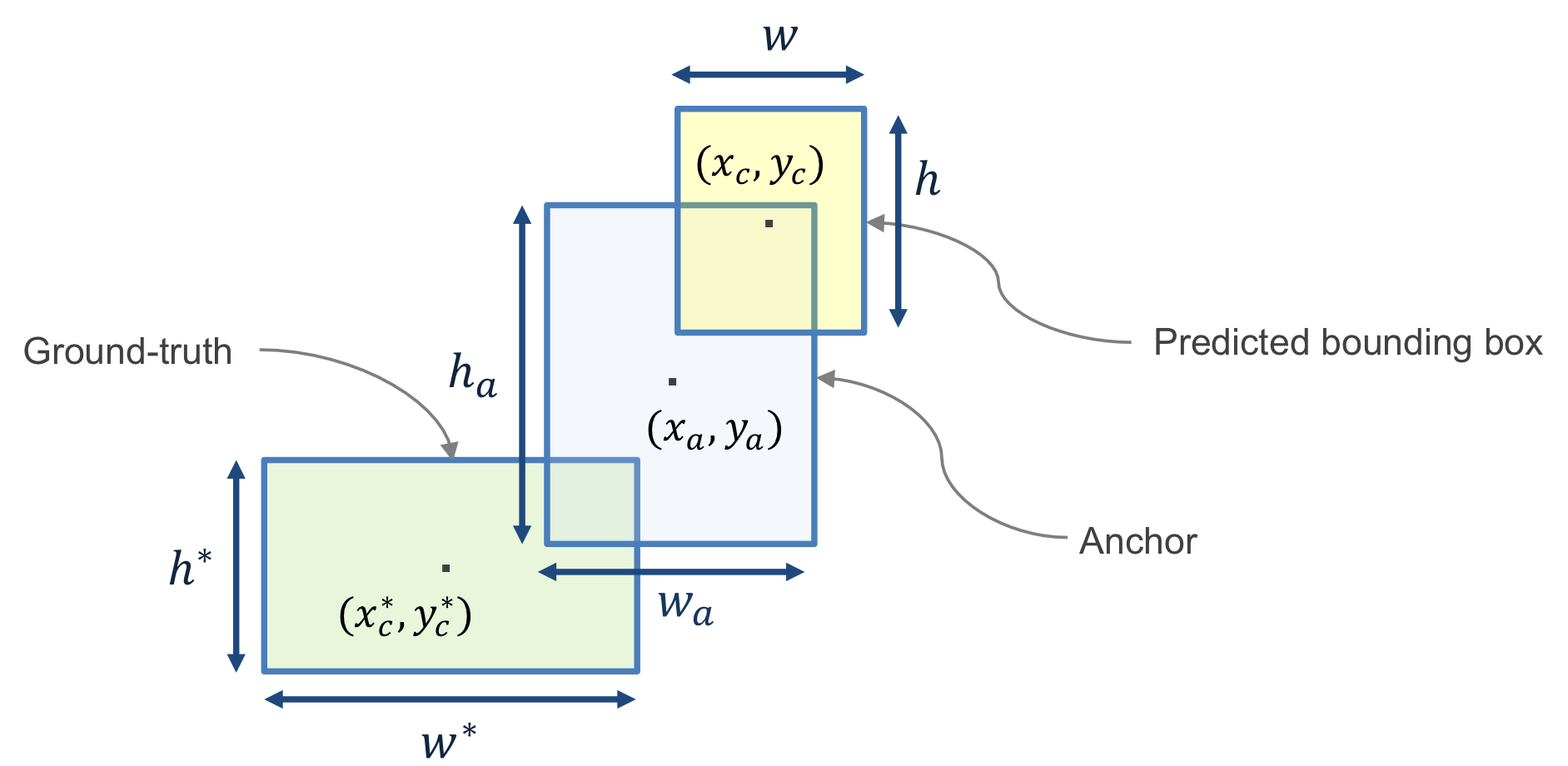}
\caption{The geometry of an anchor, a predicted bounding box, and a ground truth box.}
\label{figure9}
\end{figure}

\textbf{Training}: Training the RPN involves minimizing a combined classification and regression loss, that is now described. For each anchor, a, the best matching defect bounding box b is selected using the intersection over union (IoU) metric. If such a match is found, then a is assumed to contain a defect and it is assigned a ground-truth class label $p_a^*=1$. In this case, a vector encoding of box b with respect to anchor a is created, and denoted $\phi(b;a)$. If no match is found, then a does not contain a defect and the class label is set $p_a^*= 0$. At training time, the location loss function $L_loc$  captures the distance between the true location of a bounding box and the location of the region proposal \cite{ren_faster_2015}. The location-based loss for a is expressed as a function of the predicted box encoding $f_loc (\mathcal{I};a,\boldsymbol{\theta})$ and ground truth $\phi(b_a;a)$:\\
\begin{align}
L_{loc}(a,\mathcal{I};\boldsymbol{\theta})= p_a^* \cdot \ell_{\text{smoothL1}}\big( \phi(b_a;a)-f_{loc} (\mathcal{I};a,\boldsymbol{\theta})\big),
\end{align}\\
where $\mathcal{I}$ is the image, $\boldsymbol{\theta}$ is the model parameters, and $\ell_{\text{smoothL1}}$ is the smooth L1 loss function, as defined in \cite{girshick_fast_2015}. The box encoding of box $b$ with respect to $a$ is a vector:\\
\begin{align}
\phi(b;a)= \bigg[ \frac{x_c}{w_a} ,\frac{y_c}{h_a} ,\log ⁡w,\log ⁡h \bigg]^T,
\end{align} \\
where $x_c$ and $y_c$ are the center coordinates of the box, $w$ is the box width, and h is the box height. $w_a$ and $h_a$ are the width and height of the anchor $a$. The geometry of an anchor, $a$ predicted bounding box, and a ground truth box is shown diagrammatically in Figure \ref{figure9}. The classification loss is expressed as a function of the predicted class $f_{cls} (I;a,\boldsymbol{\theta})$ and $p_a^*$:
\begin{align}
L_{cls} (a,\mathcal{I};\boldsymbol{\theta}) = L_{CE} (p_a^*,f_{cls}(\mathcal{I};a,\boldsymbol{\theta})),
\end{align} \\
where $L_{CE}$ is the cross-entropy loss function. The total loss for $a$ is expressed as the weighted sum of the location-based loss and the classification loss \cite{huang_speed/accuracy_2017}:
\begin{align}
L(\mathcal{I};\boldsymbol{\theta})=\alpha \cdot L_{loc} (a,\mathcal{I};\boldsymbol{\theta})  + \beta \cdot L_{cls} (a,\mathcal{I};\boldsymbol{\theta}),
\end{align} \\
where $\alpha$, $\beta$ are weights chosen to balance localization and classification losses \cite{huang_speed/accuracy_2017}. To train the object detection model, (5) is averaged over the set of anchors and minimized with respect to parameters $\boldsymbol{\theta}$.

\begin{figure}[tb]
\centering
\includegraphics[width=0.38\textwidth]{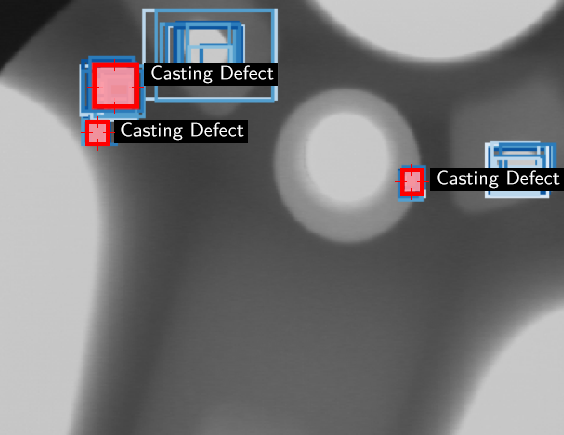}
\caption{The top 50 regions of interest, as predicted by a region proposal network trained on the Microsoft Common Objects in Context dataset. The predicted regions of interest are shown in blue, and the ground-truth casting defect locations are shown in red.}
\label{figure10}
\end{figure}

\textbf{Transfer Learning}:
The RPN is an ideal candidate for the application of transfer learning, as it identifies regions of interest (RoIs) in images, rather than identifying particular types of objects. Transfer learning is a machine learning technique where information that is learned in one setting is exploited to improve generalization in another setting. It has been shown that transfer learning is particularly applicable for domain-specific tasks with limited training data \cite{kolar_transfer_2018,gao_deep_nodate}. When training an object detection network on a large dataset with many classes, the RPN learns to identify subsections of the image that likely contain an object, without discriminating by object class. This property is leveraged by first pretraining the object detection system on a large dataset with many classes of objects, namely the Microsoft Common Objects in Context (COCO) dataset \cite{lin_microsoft_2014}. Interestingly, when the RPN from the trained object detection system is applied to an X-ray image, it immediately identifies casting defects amongst other interesting regions of the image. The output of the RPN after training solely on the COCO dataset is shown in Figure \ref{figure10}.

\subsection{Region-Based Detector}
Thus far the defect detection system is able to select a fixed number of region proposals from the original image. This section describes how a region-based detector (RBD) is used to classify the casting defects in each region, and fine-tune the bounding box coordinates. The RBD is based on the Faster R-CNN object detection network \cite{ren_faster_2015}.

The input to the RBD is cropped from the output of ResNet-101 feature extractor, according to the shape of the regressed bounding box. Unfortunately, the size of the input is dependent on the size of the bounding box. To address this issue, an RoIAlign layer is used to convert the input to a fixed-length feature vector \cite{he_mask_2017}. RoIAlign works by dividing the $h \times w$ RoI window into an $H \times W$ grid of sub-windows of size $h/H \times w/W$. Bilinear interpolation \cite{jaderberg_spatial_2015} is used to compute the exact values of the input features at four regularly sampled locations in each sub-window. The reader is referred to \cite{girshick_fast_2015} for a more detailed description of the RoIAlign layer. The resulting feature vector has spatial dimensions $H \times W$, regardless of the input size.

Each feature vector from the RoIAlign layer is fed into a sequence of convolutional and fully connected layers. In the proposed defect detection system, the RBD contains two convolutional layers and two fully connected layers. The last fully connected layer produces two output vectors: The first vector contains probability estimates for each of the $K$ object classes plus a catch-all “background” class. The second vector encodes refined bounding-box positions for one of the $K$ classes. 
The RBD is trained by minimizing a joint regression and classification loss function, similar to the one used for the RPN. The reader is referred to \cite{girshick_fast_2015} for a detailed description of the loss function and training process. The output of the RBD for a single image is shown in Figure \ref{figure10}.

\begin{figure}[tb]
\centering
\includegraphics[width=0.48\textwidth]{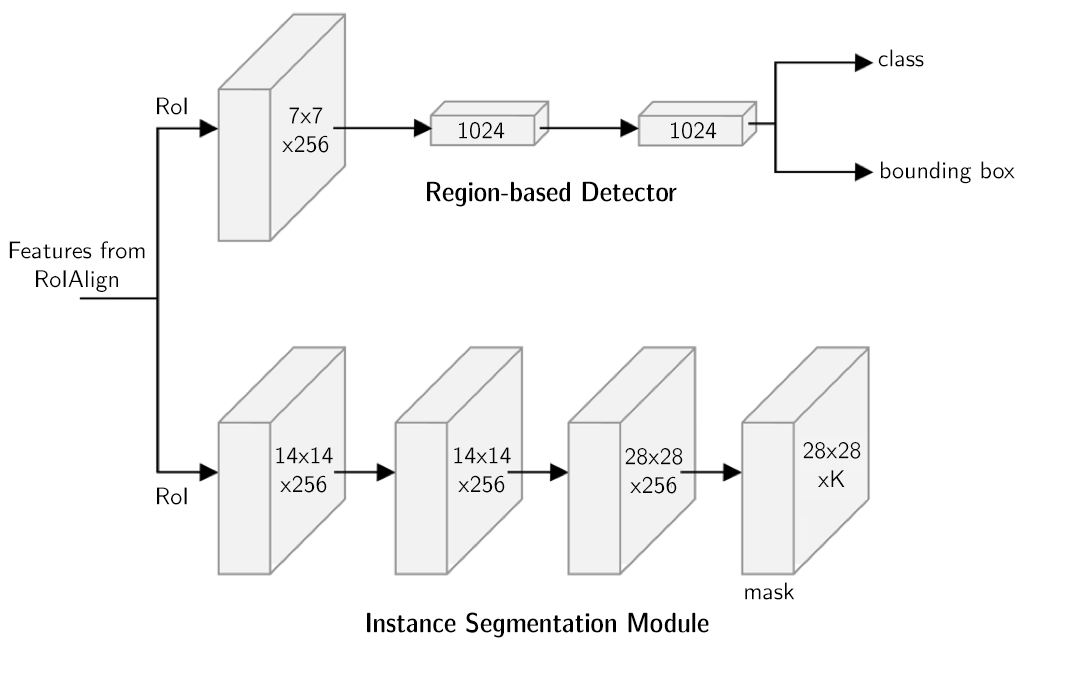}
\caption{Head architecture of the proposed defect detection network. Numbers denote spatial resolution and channels. Arrows denote either convolution, deconvolution, or fully connected layers as can be inferred from context (convolution preserves spatial dimension while deconvolution increases it). All convolution layers are $3 \times 3$, except the output convolution layer which is $1 \times 1$. Deconvolution layers are $2 \times 2$ with stride 2.}
\label{figure11}
\end{figure}

\textbf{Defect Segmentation:}
Instance segmentation is performed by predicting a segmentation mask for each RoI. The prediction of segmentation masks is performed using another CNN, referred to as the instance segmentation network. The input to the segmentation network is a block of features cropped from the output of the feature extractor. The instance segmentation network has a $28\times 28\times K$ dimensional output for each RoI, which encodes $M$ binary masks of resolution $28\times 28$, one for each of the $K$ classes. The instance segmentation network is shown alongside the RBD in Figure \ref{figure11}.

\begin{figure*}[thpb]
\centering
\includegraphics[width=0.80\textwidth]{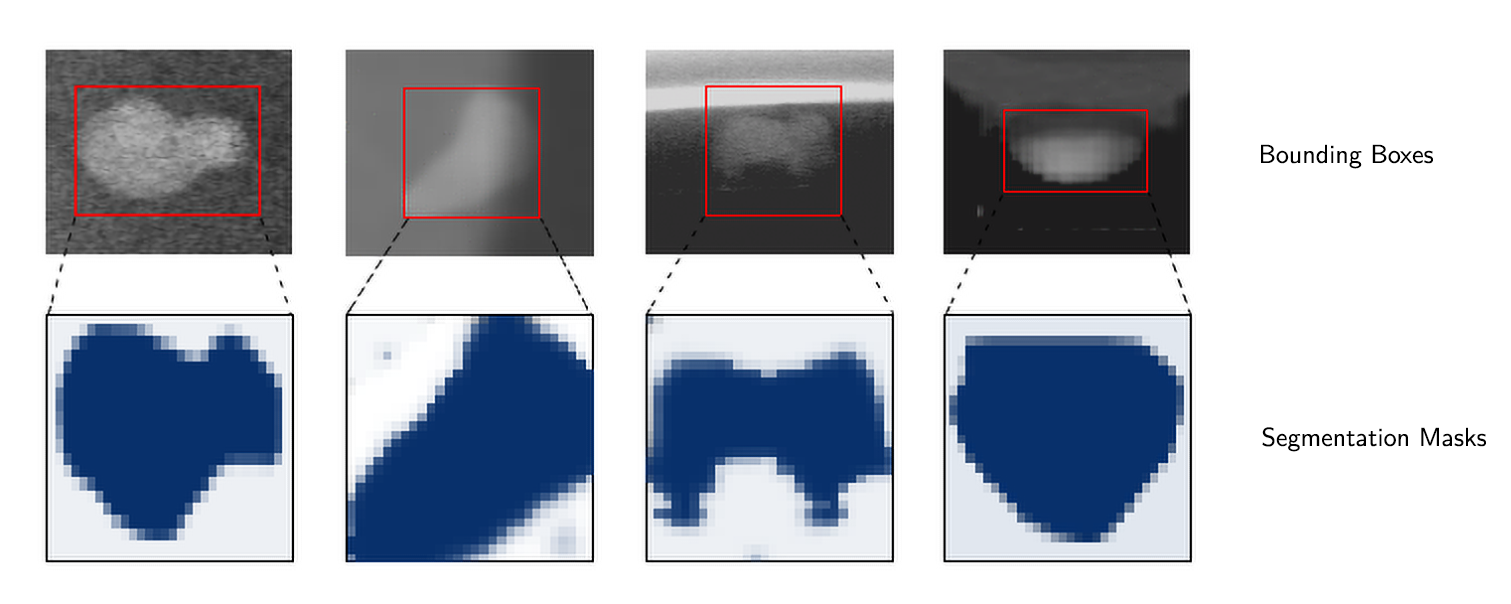}
\caption{Examples of floating point masks. The top row shows predicted bounding boxes, and the bottom row shows the corresponding predicted segmentation masks. Masks are shown here at 28 $\times$ 28 pixel resolution, as predicted by the instance segmentation module. In the proposed defect detection system, the segmentation masks are resized to the shape and size of the predicted bounding box.}
\label{figure12}
\end{figure*}

During training, a per-pixel sigmoid function is applied to the output of the instance segmentation network. The loss function $L_{mask}$ is defined as the average binary cross-entropy loss. For an RoI associated with ground-truth class $i$, $L_{mask}$ is only defined on the $i$-th mask (other mask outputs do not contribute to the loss). This definition of $L_{mask}$ allows the network to generate masks for every class without competition among classes. It follows that the instance segmentation network can be trained by minimizing the joint RBD and mask loss. At test time, one mask is predicted for each class ($K$ masks in total). However, only the $i$-th mask is used, where $i$ is the predicted class by the classification branch of the RBD. The $28 \times 28$ floating-number mask output is then resized to the RoI size, and binarized at a threshold of $0.5$. Some example masks are shown in Figure \ref{figure12}. 

\begin{figure*}[tb]
\centering
\includegraphics[width=0.70\textwidth]{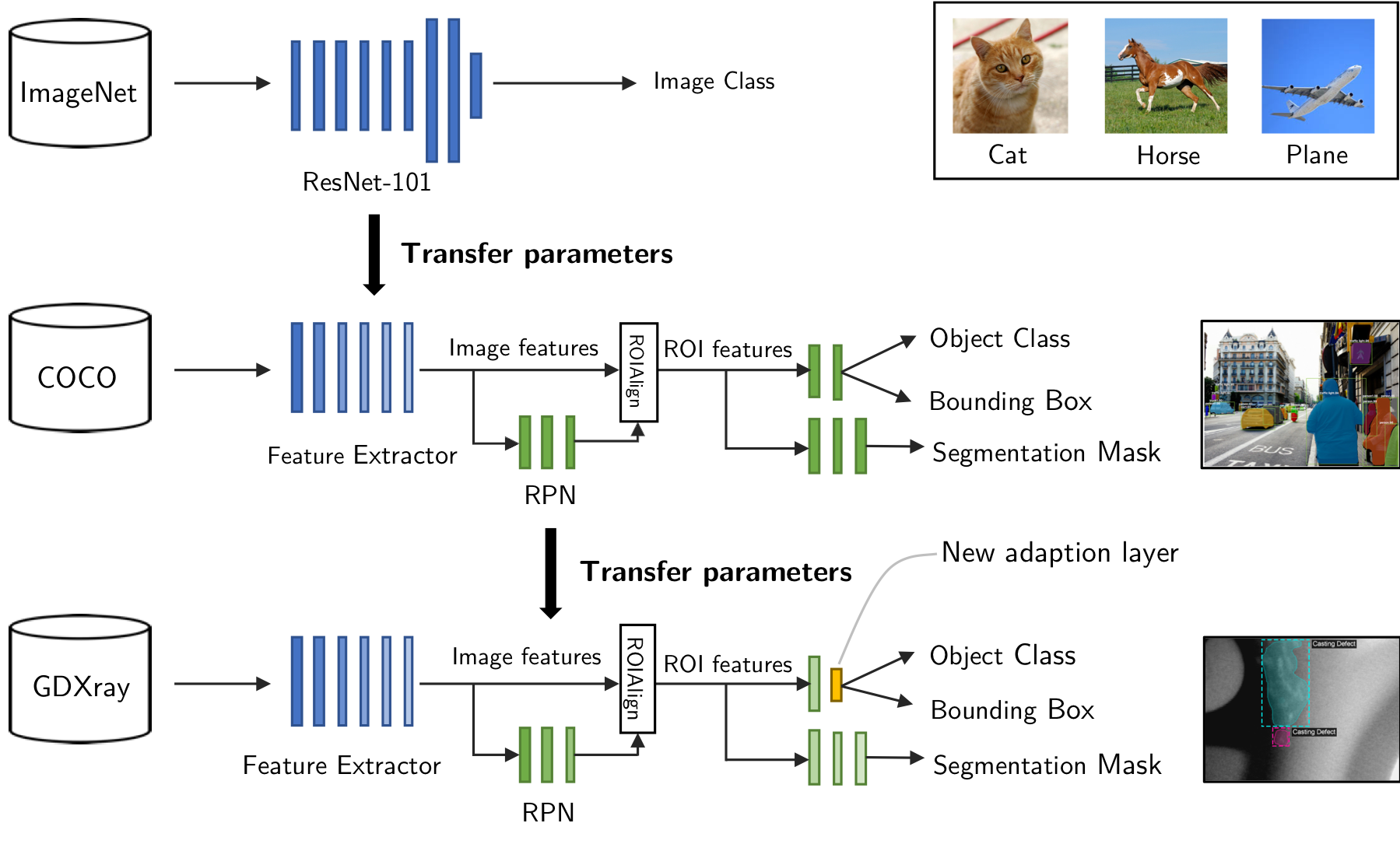}
\caption{Training the proposed defect detection system with GDXray and transfer learning. }
\label{figure13}
\end{figure*}

\section{IMPLEMENTANTION DETAILS AND EXPERIMENTAL RESULTS}
This section describes the implementation of the casting defect detection system described in the previous section.  The model is primarily trained and evaluated using images from the GDXray dataset \cite{mery_gdxray:_2015}. The Castings series of this dataset contains 2727 X-ray images mainly from automotive parts, including aluminum wheels and knuckles. The casting defects in each image are labelled with tight fitting bounding-boxes. The size of the images in the dataset ranges from $256 \times 256$ pixels to $768 \times 572$ pixels. To ensure the results are consistent with previous work, the training and testing data is divided in the same way as described in \cite{ferguson_automatic_2017}.

\subsection{Training}
The model is trained in a manner similar to many other modern object detection networks, such as Faster R-CNN and Mask R-CNN \cite{ren_faster_2015,he_mask_2017}. However, several adjustments are made to account for the small size of casting defects, and the limited number of images in the GDXray dataset. Images are scaled so that the longest edge is no larger than 768 pixels. Images are then padded with black pixels to a size of $768 \times 768$ pixels. Additionally, the images are randomly flipped horizontally and vertically at training time. No other form of preprocessing is applied to the images at training or testing time. 

Transfer learning is used to reduce the total training time and improve the accuracy of the trained models, as depicted in Figure \ref{figure13}. The ResNet-101 feature extractor is initialized using weights from a ResNet-101 network that was trained on the ImageNet dataset. The defect detection system is then trained on the COCO dataset \cite{lin_microsoft_2014}. When pretraining the model, the learning rates are adjusted to the schedule outlined in \cite{huang_speed/accuracy_2017}. Training on the relatively large COCO dataset ensures that each model is initialized to localize common objects before it is trained to localize defects. Training on the COCO dataset is conducted using 8 NVIDIA K80 GPUs. Each mini-batch has 2 images per GPU and each image has 100 sampled RoIs, with a ratio of 1:3 of positive to negatives. As in Faster R-CNN, an RoI is considered positive if it has IoU with a ground-truth box of at least 0.5 and negative otherwise.

The defect detection system is then fine-tuned on the GDXray dataset as follows: The output layers of the RBD and instance segmentation layers are resized, as they return predictions for the 80 object classes in the COCO dataset. More specifically, the output shape of these layers is resized to accommodate for two output classes, namely “Casting Defect” and “Background”. The weights of the resized layers are initialized randomly using a Gaussian distribution with zero mean and a 0.01 standard deviation. The defect detection system is trained on the GDXray dataset for 80 epochs, holding all parameters fixed except those of the output layers. The defect detection system is then trained further for an additional 80 epochs, without holding any weights fixed. 

\subsection{Inference}
The defect detection system is evaluated on a 3.6 GHz Intel Xeon E5 desktop computer machine with 8 CPU cores, 32 GB RAM, and a single NVIDIA GTX 1080 Ti Graphics Processing Unit (GPU). The models are evaluated with the GPU being enabled and disabled. For each image, the top 600 region proposals are selected by objectness score from the RPN and evaluated using the RBD. Masks are only predicted for the top 100 bounding boxes from the RBD. The proposed defect detection system is trained with and without the instance segmentation module, to investigate whether the inclusion of the instance segmentation module changes bounding box prediction accuracy. The accuracy of the system is evaluated using the GDXray Castings dataset. Every image in the testing data set is processed individually (no batching). The accuracy of each model is evaluated using the mean of average precision (mAP) as a metric \cite{manning_introduction_2008}. The IoU metric is used to determine whether a bounding box prediction is to be considered correct. To be considered a correct detection, the area of overlap $a_{o}$ between the predicted bounding box $B_{p}$ and ground truth bounding box $B_{gt}$ must exceed $0.5$ according to the formula:

\begin{align}
a_o = \frac{\text{area}(B_p  \cap B_{gt})}{\text{area}(B_{p} \cup B_
{gt})}, 
\end{align}
where $B_{p}\cap B_{gt}$ denotes the intersection of the predicted and ground truth bounding boxes and $B_{p} \cup B_{gt}$ denotes their union. The average precision is reported for both the bounding box prediction ($\text{mAP}_{bbox}$) and segmentation mask prediction ($\text{mAP}_{mask}$).

\subsection{Main Results}

\begin{figure*}[tb]
\centering
\includegraphics[width=0.90\textwidth]{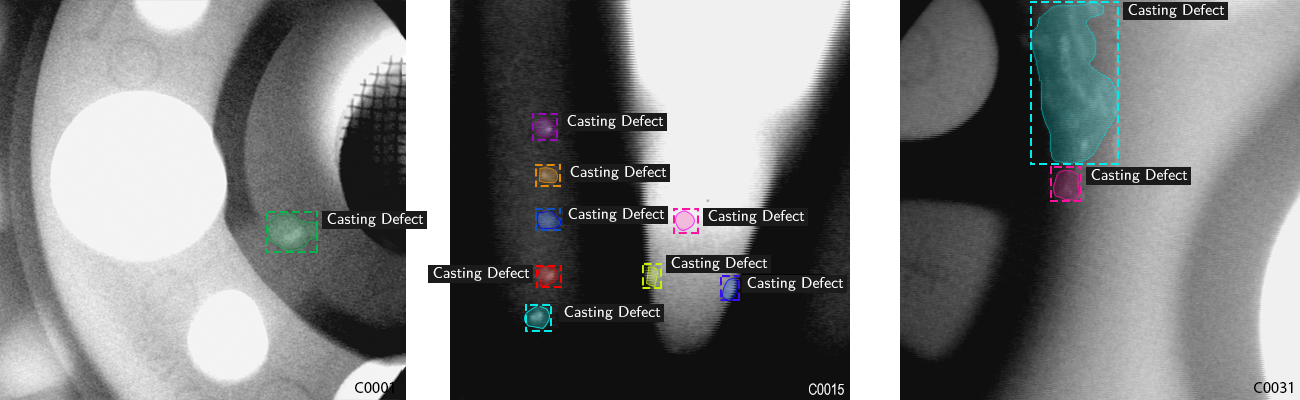}
\caption{Example detections of casting defects from the proposed defect detection system. }
\label{figure14}
\end{figure*}

As shown in Table 2, the speed and accuracy of the defect detection system is compared to similar systems from previous research \cite{ferguson_automatic_2017}. The proposed defect detection system exceeds the previous state-of-the-art performance on casting defect detection reaching an $\text{mAP}_{bbox}$ of $0.957$. Some example outputs from the trained defect detection system are shown in \ref{figure14}.  The proposed defect detection system exceeds the Faster R-CNN model from \cite{ferguson_automatic_2017} in terms of accuracy and evaluation time. The improvement in accuracy is thought to be largely due to benefits arising from joint prediction of bounding boxes and segmentation masks. Both systems take a similar amount of time to evaluate on the CPU, but the proposed system is faster than the Faster R-CNN system when evaluated on a GPU. This difference arises probably because our implementation of Mask R-CNN is more efficient at leveraging the parallel processing capabilities of the GPU than the Faster R-CNN implementation used in \cite{ferguson_automatic_2017}. It should be noted that single stage detection systems such as the SSD ResNet-101 system proposed in \cite{ferguson_automatic_2017} have a significantly faster evaluation time than the defect detection system proposed in this article.

When the proposed defect detection system is trained without the segmentation module, the system only reaches an $\text{mAP}_{bbox}$ of 0.931. That is, the bounding-box prediction accuracy of the proposed defect detection system is higher when the system is trained simultaneously on casting defect detection and casting defect instance segmentation tasks. This is a common benefit of multi-task learning which is well-documented in the literature \cite{he_mask_2017,ren_faster_2015,girshick_fast_2015}. The accuracy is improved when both tasks are learned in parallel, as the bounding box and segmentation modules use a shared representation of the input image (from the feature extractor) \cite{caruana_multitask_1997}. However, it should be noted that the proposed system is approximately 12 \% slower when simultaneously performing object detection and image segmentation. The memory requirements at training and testing time are also higher, when object detection and instance segmentation are performed simultaneously compared to pure object detection. For inference, the GPU memory requirement for simultaneous object detection and instance segmentation is 9.72 Gigabytes, which is 9 \% higher than that for object detection alone.

\begin{table*}[t]
\caption{Comparison of the accuracy and performance of each model on the defect detection task. Results are compared to the previous state-of-the-art results, presented in \cite{ferguson_automatic_2017}. }
\label{results}
\begin{center}
\setlength{\tabcolsep}{10pt}
\renewcommand{\arraystretch}{1.4}
\begin{tabular}{|m{35mm}|m{25mm}|m{25mm}|m{20mm}|m{20mm}|}
\hline
\textbf{Method} & \textbf{Evaluation time per \newline
image using CPU [s]} & \textbf{Evaluation time per \newline
image using GPU [s]} & $ \text{\textbf{mAP}}_{bbox}$ & ${\text{\textbf{mAP}}_{bbox}}$\\
\hline
Defect detection system \newline
(Object detection only) 
 & 5.340 &  0.145 & 0.931 & - \\
\hline
Defect detection system  \newline
(Detection \& segmentation) & 6.240 &  0.165 & \textbf{0.957} & \textbf{0.930} \\
\hline
Faster RCNN  & 6.319 &  0.512 & 0.921 & - \\
\hline
SSD ResNet101 & \textbf{0.141} & \textbf{0.051} & \textbf{0.762} & - \\
\hline
\end{tabular}
\end{center}
\end{table*}

\subsection{Error Analysis}
\begin{figure}[tb]
\centering
\includegraphics[width=0.5\textwidth]{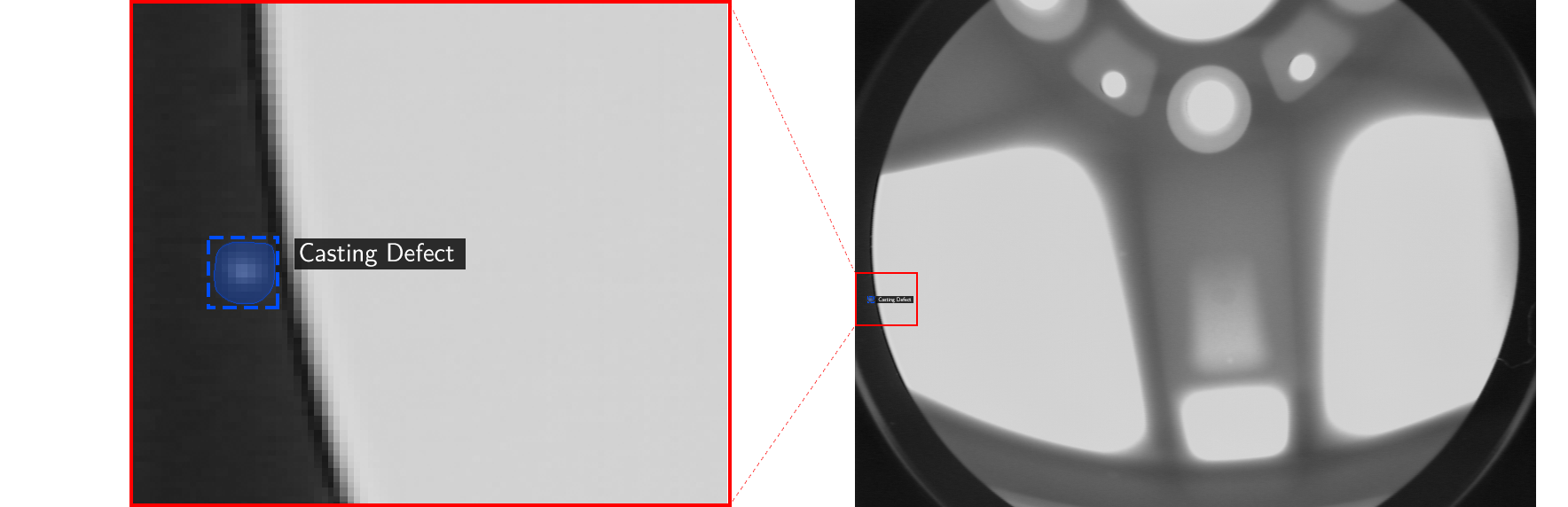}
\caption{An example of a false positive casting defect label, where a casting defect is incorrectly detected in the X-ray machine itself. This label is considered a false positive as ground-truth defects should only be labeled on the object being scanned.}
\label{figure15}
\end{figure}

\begin{figure}[tb]
\centering
\includegraphics[width=0.5\textwidth]{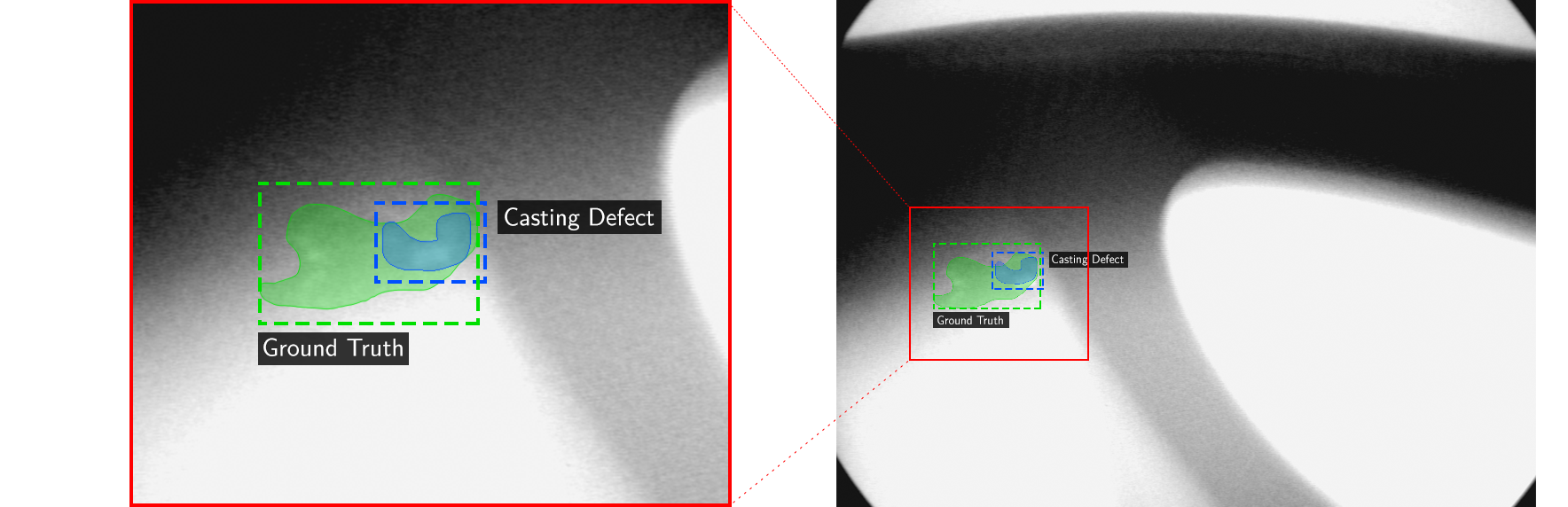}
\caption{Casting defect misclassification due to bounding box regression error. In this instance, the defect detection system failed to regress the correct bounding box coordinates resulting in a misclassification according to the IoU metric.}
\label{figure16}
\end{figure}

The proposed system makes very few misclassifications on GDXray Castings test dataset. In this section two example misclassifications are presented and discussed. Figure \ref{figure15} provides an example where the defect detection system produces a false positive detection. In this case, the proposed defect detection system identifies a region of the X-ray image which appears to be a defect in the X-ray machine itself. This defect is not included GDXray castings dataset, and hence is labelled as a misclassification.  Similar errors could be avoided in future systems by removing bounding box predictions which lie outside the object being imaged. Figure \ref{figure16} provides an example where the bounding box coordinates are incorrectly predicted, resulting in a misclassification according to the IoU metric. However, it should be noted that the label in this case is particularly subjective; the ground truth could alternatively be labelled as two small defects rather than one large one.

\section{Discussion}
During the development of the proposed casting defect detection system, a number of experiments were conducted to better understand the system. This section presents the results of these experiments, and discusses the properties of the proposed system. 

\subsection{Speed / Accuracy Tradeoff}
\begin{figure}[tb]
\centering
\includegraphics[width=0.5\textwidth]{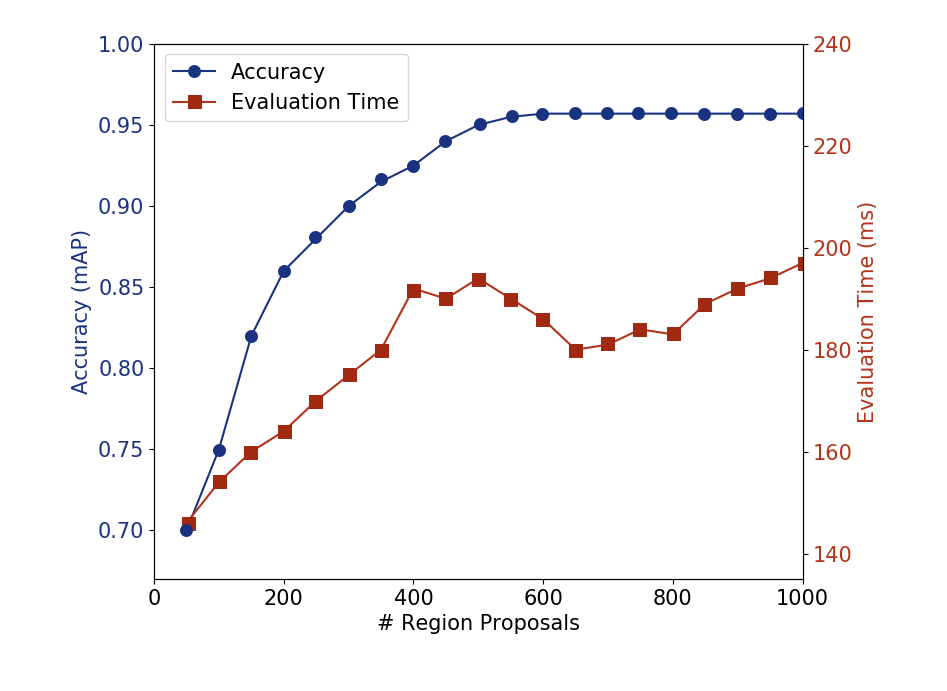}
\caption{Relationship between casting defect detection accuracy, evaluation speed, and the number of region proposals.}
\label{figure17}
\end{figure}

There is an inherent tradeoff between speed and accuracy in most modern object detection systems \cite{huang_speed/accuracy_2017}. The number of region proposals selected for the RBD is known to affect the speed and accuracy of object detection networks based on the Faster R-CNN framework \cite{he_mask_2017,ren_faster_2015,girshick_fast_2015}. Increasing the number of region proposals decreases the chance that an object will be missed, but it increases the computational demand when evaluating the network. Researchers typically achieve good results on complex object detection tasks using 3000 region proposals. A number of tests were conducted to find a suitable number of region proposals for the defect detection task. Figure \ref{figure17} shows the relationship between accuracy, evaluation time and the number of region proposals. Based on these results, the use of 600 region proposals is considered to provide a good balance between speed and accuracy.

\begin{figure}[tb]
\centering
\includegraphics[width=0.45\textwidth]{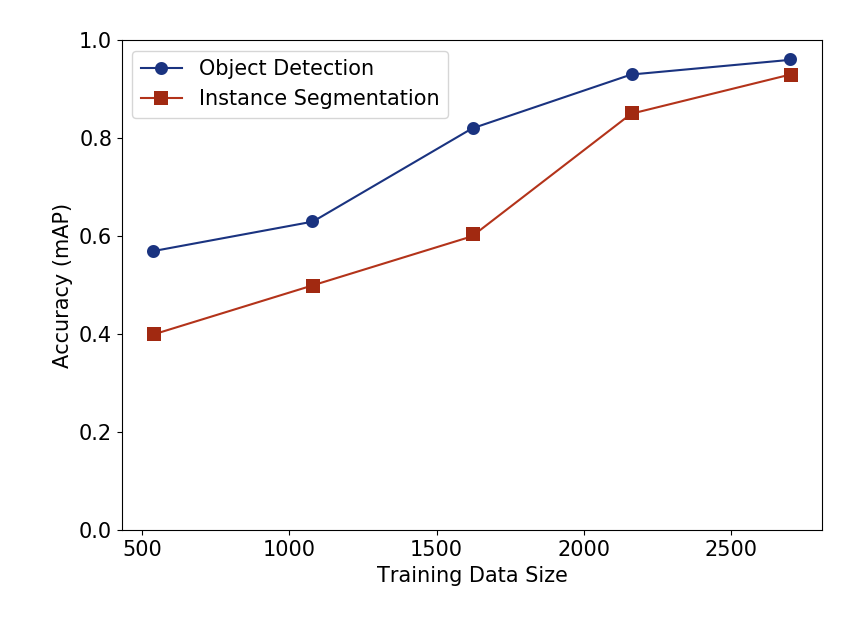}
\caption{Mean average precision (mAP) on the test set, given different sized training sets. The object detection accuracy (mAP{bbox}) and segmentation accuracy (mAP{mask}) are both shown.}
\label{figure18}
\end{figure}

\subsection{Data Requirements}
As with many deep learning tasks, it takes a large amount of labelled data to train an accurate classifier. To evaluate how the size of the training dataset influences the model accuracy, the defect detection system is trained several times, each time with a different amount of training data. The $\text{mAP}_{bbox}$ and $\text{mAP}_{mask}$ performance of each trained system is observed. Figure \ref{figure18} shows how the amount of training data affects the accuracy of the trained defect detection system. The object detection accuracy ($\text{mAP}_{bbox}$) and segmentation accuracy $\text{mAP}_{mask}$ improve significantly when the size of the training dataset is increased from $\sim$1100 to 2308 images. It also appears that a large amount of training data is required to obtain satisfactory instance segmentation performance compared to defect detection performance. Extrapolating from Figure \ref{figure18} suggests that a higher mAP could be achieved with a larger training dataset.

\subsection{Training Set Augmentation}
\begin{table*}[ht]
\caption{Influence of data augmentation techniques on test accuracy. The bounding box prediction accuracy ($\text{mAP}_{bbox}$ and instance segmentation accuracy $\text{mAP}_{mask}$ are reported on the GDXray Castings test set. }
\label{augmentation}
\begin{center}
\setlength{\tabcolsep}{10pt}
\renewcommand{\arraystretch}{1.4}
\begin{tabular}{|m{20mm} m{20mm} m{20mm} m{20mm} m{20mm}|m{10mm} m{10mm}|}
\hline
\textbf{Horizontal Flip} & \textbf{Vertical Flip} & \textbf{Gaussian Blur} &  \textbf{Gaussian Noise} & \textbf{Random \newline Cropping} & $\text{\textbf{mAP}}_{bbox}$ & $\text{\textbf{mAP}}_{mask}$ \\
\hline
- & - & - & - & - & 0.907 & 0.889\\
\hline
Yes & - & - & - & - & 0.936 & 0.920 \\
\hline
Yes & Yes &  - & - & - &0.957 & 0.930 \\
\hline
Yes & Yes & Yes & - & - &0.854 & 0.832 \\
\hline
Yes & Yes & - & Yes & - &0.897 & 0.883\\
\hline
Yes & Yes & - & - & Yes &0.950 & 0.931\\
\hline
\end{tabular}
\end{center}
\end{table*}

It is well-documented that training data augmentation can be used to artificially increase the size of training datasets, and in some cases, lead to increased prediction accuracy \cite{he_mask_2017,girshick_fast_2015}. The effect of several common image augmentation techniques on testing accuracy is evaluated in this section. Randomly horizontally flipping images is a technique where images are horizontally flipped at training time. This technique tends to be beneficial when training CNNs, as the label of an object is agnostic to horizontal flipping. On the other hand, vertical flipping is less common as many objects, such as cars and trains, seldomly appear upside-down. Gaussian blur is a common technique in image processing as it helps to reduce random noise that may have been introduced by the camera or image compression algorithm \cite{takeda_kernel_2007}. In this study, the Gaussian blur augmentation technique involved convolving each training image with a Gaussian kernel using a standard deviation of 1.0 pixels. Adding Gaussian noise to the training images is also a common technique for improving the robustness of the trained model to noise in the input images \cite{zheng_improving_2016}. In this study, zero-mean Gaussian noise with a standard deviation equal to 0.05 of the image dynamic range, is added to each image. In this context, the dynamic range of the image is defined as the range between the darkest pixel and the lightest pixel in the image. The augmentation techniques are applied during the training phase only, with the original images being used at test time.

\subsection{Transfer Learning}
This study hypothesized that transfer learning is largely responsible for the high prediction accuracy obtained by the proposed defect detection system. The system is able to generate meaningful image features and good region proposals for GDXray casting images, before it is trained on the GDXray Casting dataset. This is made possible by initializing the ResNet feature extractor using weights pretrained on the ImageNet dataset and subsequently training the defect detection system on the COCO dataset. To test the influence of transfer learning, three training schemes are tested: In training scheme (a) the proposed defect detection system is trained on the GDXray Castings dataset without pretraining on the ImageNet or COCO datasets. Xavier initialization \cite{glorot_deep_2011} is used to randomly assign the initial weights to the feature extraction layers. In training scheme (b) the same training process is repeated but the feature extractor weights are initialized using weights pretrained on the ImageNet dataset. Training scheme (c) uses pretrained ImageNet weights COCO pretraining, as described in the "Defect Detection System" section.

In Table 4, each trained system is evaluated on the GDXray Castings test dataset. Training scheme (a) does not leverage transfer learning, and hence the resulting system obtains a low $\text{mAP}_{bbox}$ of 0.651 on the GDXray Castings test dataset. In training scheme (b), the feature extractor is initialized using pretrained ImageNet, and hence the system obtains a higher $\text{mAP}_{bbox}$ of 0.874 on the same dataset. By fully leveraging transfer learning, training scheme (c) leads to a system that obtains a $\text{mAP}_{bbox}$ of 0.957, as described earlier. In Table 4, the mAP of the trained systems is also reported on the GDXray Castings training dataset. In all cases, the model fits the training data closely, demonstrating that transfer learning affects the system’s ability to generalize predictions to unseen images rather than its ability to fit to the training dataset.

\begin{table*}[ht]
\caption{Quantitative results indicating the influence of transfer learning on the accuracy of the trained defect detection system. The bounding box prediction accuracy $\text{mAP}_{bbox}$ and instance segmentation accuracy 
$\text{mAP}_{mask}$ are reported on the GDXray Castings training dataset and GDXray Castings test dataset. }
\label{augmentation}
\begin{center}
\setlength{\tabcolsep}{10pt}
\renewcommand{\arraystretch}{1.5}

\begin{tabular}{|m{10mm} |m{25mm} | m{25mm} | m{20mm} m{20mm} | m{20mm} m{10mm} |}
\hline
 &  & &  \multicolumn{2}{|c|}{\textbf{GDXRay Castings Training Set}}  & \multicolumn{2}{|c|}{\textbf{GDXRay Castings Test Set}}  \\
\hline
\textbf{Training Scheme} & \textbf{Feature Extractor Initialization} & \textbf{Pretraining on MS COCO Dataset} & $\text{\textbf{mAP}}_{bbox}$ & $\text{\textbf{mAP}}_{mask}$ & $\text{\textbf{mAP}}_{bbox}$ & $\text{\textbf{mAP}}_{mask}$\\
\hline
a &Xavier Initialization \cite{glorot_understanding_2010} (Random) & No & 0.970 & 0.960 & 0.651 & 0.420 \\
\hline
b &Pretrained ImageNet Weights & No  & 1.00 & 0.981 & 0.874 & 0.721 \\
\hline
c &Pretrained ImageNet Weights & Yes & 1.00 & 0.991 & \textbf{0.957} & \textbf{0.930} \\
\hline
\end{tabular}
\end{center}
\end{table*}

\subsection{Weld defect segmentation with multi-class learning}
\begin{figure}[ht]
\centering
\includegraphics[width=0.4\textwidth]{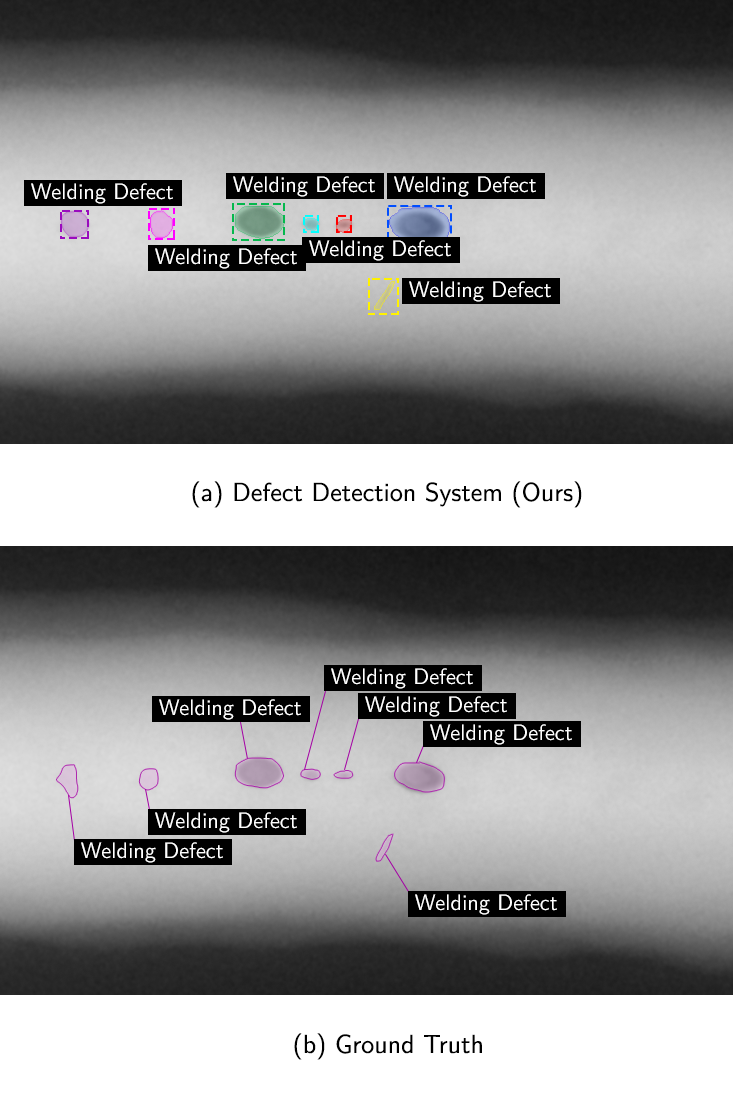}
\caption{Comparison of weld defect detections to ground truth data, using one image from the GDXray Welds series. The task is primarily an instance segmentation task, so the ground truth bounding boxes are not shown.}
\label{figure19}
\end{figure}

The ability to generalize a model to multiple tasks is highly beneficial in a number of applications. The proposed defect detection system was retrained on both the GDXray Castings dataset and the GDXray Welds dataset. The GDXray Welds dataset contains 88 annotated high-resolution X-ray images of welds, ranging from 3176 to 4998 pixels wide. Each high-resolution image is divided horizontally into 8 smaller images for testing and training, yielding a total of 704 images. 80 \% of the images are randomly assigned to the training set, with the remaining 20 \% assigned to the testing set. Unlike the GDXray Castings dataset, the GDXray Welds dataset is only annotated with segmentation masks. Bounding boxes are fitted to the segmentation masks by identifying closed shapes in the mask using a binary border following algorithm \cite{suzuki_topological_1985}, and wrapping each shape in  a tightly fitting bounding box. The defect detection system is simultaneously trained on images from the Castings and Welds training sets. The defect detection system is able to simultaneously identify casting defects and welding defects, reaching a segmentation accuracy $\text{mAP}_{mask}$ of 0.850 on the GDXray Welds test dataset. Some example predictions are shown in Figure \ref{figure19}. The detection and segmentation of welding defects can be considered very accurate, especially given the small size of the GDXray Welds dataset with only 88 high-resolution images. Unfortunately, there is no measurable improvement on the accuracy of casting defect detection when jointly training on both datasets

\subsection{Defect Detection on Other Datasets Using Zero-Shot Learning}
\begin{figure}[ht]
\centering
\includegraphics[width=0.46\textwidth]{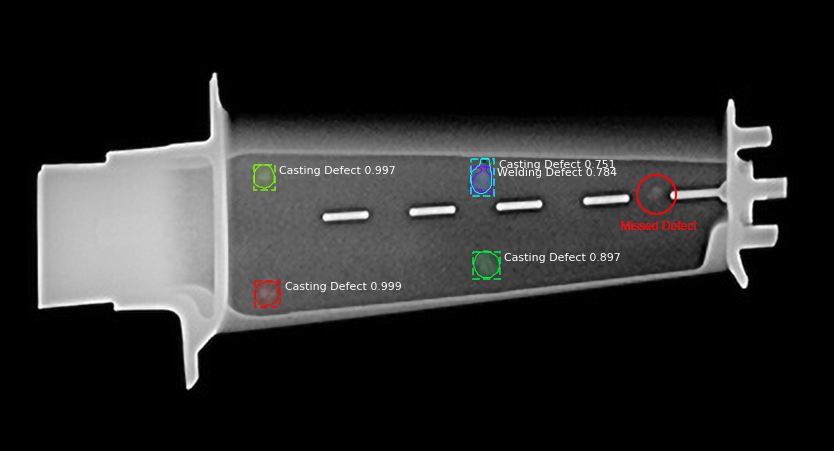}
\caption{Defect detection and segmentation results on an X-ray image of a jet turbine blade. The training set did not contain any turbine blade images. The defect detection system correctly identifies four out of the 5 five defects in the image. The top right defect is incorrectly classified as both a “Casting Defect” and a “Welding Defect”. }
\label{figure20}
\end{figure}

A good defect detection system should be able to classify defects for a wide range of different objects. The defect detection system can be said to generalize well if it is able to detect defects in objects that do not appear in the training dataset. In the field of machine learning, zero-shot transfer is the process of taking a trained model, and using it, without retraining, to make predictions on an entirely different dataset. To test the generalization properties of the proposed defect detection system, the trained system is tested on a range of X-ray images from other sources. The system correctly identifies a number of defects in a previously unseen X-ray image of a jet turbine blade, as shown in Figure \ref{figure20}. The jet turbine blade contains five casting defects, of which four are identified correctly. It is unsurprising that the system fails to identify one of the casting defects in the image, as there are no jet engine turbine blades in the GDXray dataset. Nonetheless, the fact that the system can identify defects in images from different datasets demonstrates its potential for generalizability and robustness.

\section{Summary and Conclusion}
This work presents a defect detection system for simultaneous detection and segmentation of defects in metal castings. This ability to simultaneously perform defect detection and segmentation makes the proposed system suitable for a range of automated quality control applications. The proposed defect detection system exceeds state-of-the-art performance for defect detection on the GDXray Castings dataset obtaining a mean average precision ($\text{mAP}_{bbox}$) of 0.957, and establishes a new benchmark for instance segmentation on the same dataset. This high-accuracy system is developed by leveraging a number of powerful paradigms in machine learning, including transfer learning, dataset augmentation, and multi-task learning. The benefit of the application of each of these paradigms was evaluated quantitatively through extensive ablation testing.

The defect detection system described in this work is able to detect casting and welding defects with very high accuracy. Future work could involve training the same network to detect defects in other materials such as wood or glass. The proposed defect detection system was designed for multi-class detection, so the system could naturally be extended detect a range of different defect types in multiple materials. The defect detection system described in this work could also be trained to detect defects in additive manufacturing applications. \\
The proposed defect detection system is accurate and performant enough to be useful in a real manufacturing setting. However, the training process for the system is complex and computationally expensive. Future work could focus on developing a standardized method of representing these models, making it easier to distribute the trained models. 

\section*{Acknowledgements}

The authors acknowledge the support by the Smart Manufacturing Systems Design and Analysis Program at the National Institute of Standards and Technology (NIST), US Department of Commerce.  This work was performed under the financial assistance award (NIST Cooperative Agreement 70NANB17H031) to Stanford University. Certain commercial systems are identified in this article. Such identification does not imply recommendation or endorsement by NIST; nor does it imply that the products identified are necessarily the best available for the purpose. Further, any opinions, findings, conclusions, or recommendations expressed in this material are those of the authors and do not necessarily reflect the views of NIST or any other supporting U.S. government or corporate organizations. 

\bibliography{library} 
\bibliographystyle{ieeetr}
\end{document}